\documentclass{new_tlp}
\usepackage{mathptmx}

\usepackage{graphicx}
\usepackage{url}
\usepackage{booktabs}
\usepackage{subfig}
\usepackage{amssymb}
\usepackage{amsfonts}
\usepackage{amsmath}
\usepackage{todonotes}
\usepackage{tcolorbox}
\usepackage{xspace}
\usepackage{paralist}
\usepackage[inline,shortlabels]{enumitem}

\usepackage{color} 
\newcommand{\red}{black}

\newcommand{\nop}[1]{}
\newcommand{\naf}{\ensuremath{\mathtt{not\ }}}
\newcommand{\R}{\ensuremath{r}}
\newcommand{\derives}{\ensuremath{\mbox{\,$:$--}\,}\xspace}
\newcommand{\Or}{\ensuremath{|}\xspace}
\newcommand{\p}{\ensuremath{{\cal P}}\xspace}

\newcommand{\BP}{\ensuremath{B_{\p}}\xspace}
\newcommand{\UP}{\ensuremath{U_{\p}}\xspace}
\newcommand{\BpR}{\ensuremath{B^+(\R)}}
\newcommand{\BnR}{\ensuremath{B^-(\R)}}
\newcommand{\mm}[1]{\mbox{${\rm MM}(#1)$}}
\newcommand{\additionalmaterial}{\url{https://www.mat.unical.it/calimeri/material/mix-lp-nn/}}

\title[Theory and Practice of Logic Programming]
        {{\color{\red}A Logic-Based Framework Leveraging \\ Neural Networks for Studying the Evolution of \\ Neurological Disorders}}

\author[F. Calimeri et al.]
{
FRANCESCO CALIMERI, FRANCESCO CAUTERUCCIO, LUCA CINELLI\\
DEMACS, University of Calabria, Italy\\
\email{\{calimeri,cauteruccio,cinelli\}@mat.unical.it}
\and
ALDO MARZULLO\\
DEMACS, University of Calabria, Italy - \\
CREATIS; CNRS UMR5220; INSERM U1206; Universit\'{e} de Lyon, Universit\'{e} Lyon~1, INSA-Lyon, Villeurbanne, France\\
\email{marzullo@mat.unical.it}
\and
CLAUDIO STAMILE\\
CREATIS; CNRS UMR5220; INSERM U1206; Universit\'{e} de Lyon, Universit\'{e} Lyon~1, INSA-Lyon, Villeurbanne, France\\
\email{stamile@creatis.insa-lyon.fr}
\and
GIORGIO TERRACINA\\
DEMACS, University of Calabria, Italy\\
\email{terracina@mat.unical.it}
\and
FRAN\c{C}OISE DURAND-DUBIEF\\
H\^{o}pital Neurologique, Service de Neurologie A\\ Hospices Civils de Lyon, Bron - France - \\
CREATIS; CNRS UMR5220; INSERM U1206; Universit\'{e} de Lyon, Universit\'{e} Lyon~1, INSA-Lyon, Villeurbanne, France\\
\email{francoise.durand-dubief@chu-lyon.fr}
\and
DOMINIQUE SAPPEY-MARINIER\\
CERMEP - Imagerie du Vivant;\\ Universit\'{e} de Lyon, Bron, France - \\
CREATIS; CNRS UMR5220; INSERM U1206; Universit\'{e} de Lyon, Universit\'{e} Lyon~1, INSA-Lyon, Villeurbanne, France\\
\email{dominique.sappey-marinier@univ-lyon1.fr}
}

\jdate{?}
\pubyear{?}
\pagerange{\pageref{firstpage}--\pageref{lastpage}}
\doi{?}

\newtheorem{example}{Example}[section]

\begin{document}

\label{firstpage}

\maketitle

\begin{abstract}
Deductive formalisms have been strongly developed in recent years; {\color{\red} among them, Answer Set Programming (ASP) gained some momentum, and has been lately fruitfully employed in many real-world scenarios.}
Nonetheless, in spite of a large number of success stories in relevant application areas, and even in industrial contexts, deductive reasoning {\color{\red} cannot} be considered the ultimate, comprehensive solution to AI; indeed, in several contexts, other approaches {\color{\red} result to be more useful}.
Typical Bioinformatics tasks, {\color{\red} for instance} classification, are currently carried out mostly by Machine Learning (ML) based solutions.

{\color{\red} In this paper, we focus on the relatively new problem of analyzing  the {\em evolution} of neurological disorders.
In this context, ML approaches already demonstrated to be a viable solution for classification tasks; here, we show how ASP can play a relevant role in the {\color{\red} brain} evolution simulation task.
In particular, we propose a general and extensible framework to support physicians and researchers at understanding the complex mechanisms underlying neurological disorders.
The framework relies on a combined use of ML and ASP, and is general enough to be applied in several other application scenarios, which are outlined in the paper. 
\newline 
Under consideration in Theory and Practice of Logic Programming (TPLP)}
\end{abstract}


\section{Introduction}
\label{sec:introduction}

After a number of works on automatic theorem proving and artificial intelligence, the work on logic programming took actually off in the 1970's, with the aim of obtaining automated deduction systems.
{\color{\red}Answer Set Programming (ASP)~\cite{gelf-lifs-91,niem-99,DBLP:journals/tplp/LoncT06}
is one of the several formalisms that stemmed out of such research efforts, and during the years it turned out to be a powerful declarative formalism for knowledge representation and reasoning (KRR).}
%
%
After more than twenty-five years of scientific research, the theoretical properties of the language are considered to be well understood, and even if the community is still very active on several extensions, the solving technology, as witnessed by the availability of a number of robust and efficient systems~\cite{DBLP:journals/jair/GebserMR17,DBLP:conf/ijcai/GebserLMPRS18},
is mature for practical applications \cite{DBLP:journals/jar/BrooksEEMR07,Terracina-TPLP1,gebser_schaub_thiele_veber_2011,Terracina-TPLP3}.
In the latest years, ASP has been indeed employed in many different domains, and used for the development of enterprise and industrial-level applications~\cite{calimeri2013application,leon-ricc-2015-rw-invited,DBLP:journals/aim/ErdemGL16}, fostered by the release of a variety of proper development tools and interoperability mechanisms for allowing interaction and integration with external systems~\cite{DBLP:conf/asp/Ricca03,DBLP:journals/amai/CalimeriCI07,febb-etal-2011,DBLP:conf/kr/FebbraroLGR12,Thimm:2014,DBLP:conf/iclp/SchullerW15,gekakasc14b,DBLP:conf/birthday/EiterRS16,DBLP:conf/padl/RathR17,DBLP:conf/aiia/CalimeriFPZ17,eggkrsw2018-ki,DBLP:journals/ngc/CalimeriFGPZ19}.


Nevertheless, even though Answer Set Programming, and deductive approaches via logic formalisms in general, are of wide use for Artificial Intelligence (AI) applications, they are not the ultimate, comprehensive solution to AI, as some kind of problems can be hardly encoded by logic rules.
This can be due to several reasons: the nature of the problem, the lack of proper development tools, and severe performance issues even for the best performing systems.
Bioinformatics is a prime example, as both related data (think, for instance, of biomedical images, temporal data, etc.) and relevant tasks (think, for instance, of classification) are not naturally approachable with deductive strategies.
In such research area, and similar ones, different AI-based strategies have been employed.
Lately, approaches relying on Machine Learning (ML) and Artificial Neural Networks (ANNs)~\cite{Haykin:1998:NNC:521706,DBLP:books/daglib/0040158} are on the rise, both because of the great results achieved by the research community in the latest years and because they better deal with the data and the nature of the tasks of interest.

Basically, with ANNs the problem is not actually modeled, and its structure remains almost unknown; rather, approaches progressively learn, usually by examples, the best answers {\color{\red} to provide in presence of specific} inputs.
ANNs can learn and model non-linear complex relationships, and even generalize acquired knowledge in order to infer new properties over unseen data; moreover, once trained, a neural network can be extremely fast in providing answers to {\color{\red} instances of} complex problems.
Unfortunately{\color{\red},} obtained results have only statistical significance; {\color{\red}it is noteworthy, indeed, that the main weakness of ANNs is in general their incompleteness, since their precision may strongly depend on the training phase and on the quality of the training data.}
The logic they use can be sound, yet proven incorrect by further observations.
{\color{\red} Hence, c}learly, as for deductive reasoning, not all problems can be properly solved by ANNs.

{\color{\red} This paper focuses on some opportunities provided by a combined use of ASP and ANN.
As a motivating scenario, we concentrate on a relevant bioinformatics problem, namely the study of neurological disorders and, specifically, on the Multiple Sclerosis (MS) disease.
To the best of our knowledge, computer science research on this topic mainly focused on the identification of the disease and, possibly, on the detection of its severity.
Actually, in order to clinically analyze disease evolution, long periods of observations, even decades, are needed.
Obviously, a tool supporting physicians in accelerating the analysis process, e.g. via simulations, would be of great benefit.
Notably, it has been demonstrated by several independent studies that there is a strong correlation between the variations of the structure of the connections among neurons (also called {\em connectome}) with possible insurgence of several neurological disorders~\cite{BaMa13}.
Hence, it would be of high interest to simulate the course of the disease by simulating brain connections degradation, in order to understand which kind of modifications might mostly determine an evolution of the disease into a worst state, or which recovery processes might induce a remission state.
Unfortunately, this simulation represents a non-trivial challenge due to various reasons, including the fact that the actual mechanisms guiding the evolution of the pathology are still largely unknown.
%

Interestingly, the connectome can be fruitfully represented by means of a graph;
hence, a possible solution would be to simulate the progress of the pathology by means of a set of custom-defined rules for modelling the evolution of the brain structure, which may involve a certain background knowledge.
%
%
An effective {\color{\red} tool for the} experts could consist of a comprehensive environment which allows to dynamically detect minimal alterations of brain connections, based on specific guidelines, that induce a change of state in the disease.
The possible change of state can, in turn, be detected by neural networks, exploring latent relations learnt from samples.
{\color{\red} It is worth pointing out that, to the best of our knowledge, a tool providing these features is not available yet; moreover, simulating {\em manually} brain alterations is not an easy task and it is almost impossible to  manually detect significant brain substructures.
	
In this context, ASP can play a relevant role.
In fact, simulation tasks could be in principle designed with any programming paradigm; nevertheless, since it is still not clear which changes to which graph properties are likely to impact the evolution of the disease, a try-and-check methodology is necessary.
This implies writing ad-hoc simulation machineries for each of them.
Clearly, a declarative methodology allows for compact and clear definitions as well as fast prototyping: ASP paves the way to easy definition of rules for the identification of brain substructures that can be of interest for the analyst.

Interestingly, it has been shown~\cite{graphbased} that it is important to look at minimal graph changes allowing to reach a certain goal in some graph variations, such as density or assortativity; this implies that most of the tasks to be carried out would actually involve optimization.
The use of ASP for optimization problems is still relatively less popular than its use for decision problems, even though it has been proved to be perfectly suited for them; moreover,
the formulas for the computation of graph metrics can be in general not easy to be written by rules only.
This increases the interest in the application of recent extensions of ASP systems, such as I-DLV/DLV2 \cite{DBLP:journals/ia/CalimeriFPZ17,Adrian2018-KI-dlv2}, DLVHEX \cite{DBLP:journals/tplp/EiterFIKRS16,DBLP:journals/tciaig/CalimeriFGHIR0T16,eggkrsw2018-ki}, etc., which allow both to solve optimization problems and the integration of external computation sources within the ASP program.
The potentialities of ASP in these two areas are pointed out in the rest of the paper.

\medskip

The main contribution of the paper consists of a framework for studying the evolution of neurological disorders by simulating variations in the connectome.
The framework relies on two main modules: an ANN that classifies a given connectome with respect to disorder stages, and a logic program (specified via ASP) that allows to perform non-deterministic variations on a given connectome that guarantee fixing some graph parameters under study.
The framework iterates between the ANN and the ASP modules to simulate possible evolutions of a neurological disorder.

The present paper elaborates on, and extends, our previous work presented in \cite{DBLP:conf/ruleml/CalimeriCMST18} as follows:
\begin{itemize}
\item
	it presents a thorough analysis on related work;	
\item
    it extends the general framework with new modules and new functionalities;
\item
    it introduces two new actualizations of the framework, thus showing three use cases benefiting of the framework that are also biologically relevant;
\item
    it improves the ANN classifier for MS in order to provide insights on the classification activity, thus allowing a more refined combined use of ASP and ANN;
\item
    it significantly extends the experimental activity, now including the new use cases and a specific performance analysis;
\item
    it introduces a web tool implementing the proposed framework and developed to support physicians in their analyses via the framework;
\item
    it overviews two more applications possibly benefiting of specializations of our general framework.
\end{itemize}

The remainder of the paper is structured as follows.
In Section~\ref{sec:related} we discuss related work, and in Section~\ref{sec:background} we recall some preliminary notions both from the biomedical and ASP contexts.
In Section~\ref{sec:framework} we introduce the general framework and discuss the role ASP and ANN play therein, while
Section~\ref{sec:specializations} presents three actualizations of the framework.
Section~\ref{sec:experiments} describes the experiments we carried out to assess potentiality and applicability of the approach.
Section~\ref{sec:webtool} introduces the web tool we developed based on the presented work, whereas Section~\ref{sec:furtherapplication} surveys some other applications possibly benefiting of the presented framework.
Eventually, in Section~\ref{sec:conclusion}, we draw our conclusions and outline some future work.
}}

\section{Related Works}
\label{sec:related}

{\color{\red} In this section we outline related literature. 
Since, to the best of our knowledge, there is no proposal available, to date, that simulates the evolution of neurological disorders by a combined use of ASP and ANN, we concentrate our attention on works exploiting together, to different extents, declarative formalisms and ML solutions.} 

\medskip

{\color{\red}
Some works have been carried out to integrate data-driven solutions into declarative systems with the aim of increasing performance; for instance, such solutions are used for inductively choosing configurations, algorithms selection, and proper coupling of subsystems~\cite{DBLP:conf/lpnmr/GebserKKSSZ11,DBLP:journals/tplp/MarateaPR14,DBLP:conf/aiia/FuscaCZP17}.
In some proposals, see for instance SMT~\cite{DBLP:journals/jar/CokSW15,BarFT-SMTLIB,DBLP:reference/mc/BarrettT18,DBLP:journals/jar/BarrettDMOS13} or CASP~\cite{DBLP:conf/iclp/BaseliceBG05,DBLP:journals/amai/MellarkodGZ08,DBLP:journals/tplp/BalducciniL17,DBLP:journals/tplp/LierlerS17,DBLP:conf/kr/ShenL18,DBLP:journals/tplp/AriasCSM18}, the logic solver can select statements that should be checked by external theory/numerical solvers, so that the next steps carried out by the logic solver depend on the answers produced by the external ones.
Some works mix statistical analysis and ASP~\cite{DBLP:conf/datalog/Gelfond10,DBLP:conf/rr/NicklesM14,DBLP:conf/aaai/BeckDEF15}: here, the aim is to extend  logic programs with probabilistic reasoning, either by a direct integration or by embedding external predicates.}
Other approaches are related to the use of  methods that ``guide'' the reasoning, the generation of logic programs or other  optimizations~\cite{DBLP:journals/tplp/LawRB15,DBLP:journals/tplp/LawRB16,DBLP:conf/commonsense/ChabierskiRLB17,DoRi18}; most of them are still at a preliminary stage.

{\color{\red}
In the context of ASP, some proposals allow ASP systems to access external sources of computation and even value invention~\cite{DBLP:journals/amai/CalimeriCI07,DBLP:journals/tplp/Redl16,DBLP:conf/rweb/KaminskiSW17,DBLP:journals/ia/CalimeriFPZ17,DBLP:journals/ki/EiterGIKRSW18,DBLP:conf/lpnmr/AlvianoADLMR19,DBLP:journals/tplp/GebserKKS19},
making them to impact semantics computation to different extents. In particular, thanks to extended built-in constructs, it is possible to invoke external functions and define custom constraints; via such invocations, one might in principle place a call even to an ANN from an ASP program.}

In the last decades, ANNs have become {\color{\red}one of the most powerful machine learning tools} for solving complex problems. Especially in visual domains, they achieved impressive results in object detection, recognition and classification~\cite{DBLP:books/daglib/0040158}.
An ANN is typically represented as a composition of functions, each computing a nonlinear weighted combination of its input, which constitute the neural structure, and are organized in layers.
Starting from samples, the learning algorithm iteratively refines the network parameters $\theta$ in order to approximate a target function $f$.
For example, in the particular case of classification, the ANN learns how to approximate a function $y = f(x,\theta)$ which maps an input $x$ to a category whose label is $y$.
However, the approximation given by the ANN does not provide any insight on the form of $f$, meaning that there is no interpretable connection between the parameters and the target function.
This is one of the main causes for the interpretation of ANN being an open problem.

Several attempts have been made for interpreting the behavior of Neural Networks, also using declarative approaches~\cite{DBLP:journals/jzusc/ZhangZ18}, and to incorporate symbolic knowledge into neural networks, resulting in a class of networks known as knowledge-based neural networks~\cite{Towell93}. 
In the context of network interpretation, Zhang \emph{et al.} used explanatory graphs and decision trees to create interpretable rules describing convolutional neural network (CNN) features~\cite{Zhang2018b,DBLP:conf/cvpr/ZhangWZ18a}. 
Furthermore, based on a semantic And-Or representation, Zhang \emph{et al.} also proposed a method to use active question-answering to assign a semantic meaning to neural patterns in convolutional layers of a pre-trained CNN and built a model for hierarchical object understanding~\cite{Zhang2017b}. 
Hu \emph{et al.} proposed a framework using logic rules to obtain more meaningful network representations by constructing an iterative distillation method that transfers the structured information of logic rules into the weights of neural networks~\cite{Hu2016}.

\section{{\color{\red} Preliminaries}}
\label{sec:background}
{\color{\red}
In this section we report some preliminary background knowledge; we address the biomedical background first, before overviewing some basic knowledge on ASP.
}

\subsection{Background on neurological disorders}
\label{sec:neurological}
The incidence of neurological disorders is constantly growing, also because population is aging on the average in most countries; hence, the efforts to design approaches capable of determining the onset of these disorders and monitoring their course in patients are intensifying~\cite{Duun*12,Hornero*09,WiScZu06}.
Furthermore, the tools supporting neurologists in their activities are becoming more complex and sophisticated (think, for instance, of magnetic resonance imaging (MRI) or of new electroencephalograms (EEG) with $256$ electrodes, instead of the classical ones with $19$ electrodes).
These important advances foster the need for handling new data formats, like images and temporal series, that are hard to be analyzed by human experts.
In these scenarios, automatic tools for the analysis are becoming more and more essential.

In many neurological investigations, a key role is played by the connections between brain areas, that can be studied by means of Magnetic Resonance Imaging (MRI); graph theory, and specifically network analysis tools, may hence provide insights to evaluate the health state of the brain.
A challenging issue is to find suitable representations of brain areas as a network, and then proper tools for interpreting them.

\medskip

%
Multiple Sclerosis (MS) is a chronic disease of the central nervous system that disrupts the flow of information within the brain, and between brain and body.
In 2015, about $2.3$ million people resulted as affected, globally, with rates varying widely among different regions and different populations~\cite{msstats}.
In the majority of cases (with a probability of $85$\%), the disease starts  with a first acute episode, called Clinically Isolated Syndrome (CIS), that evolves into the Relapsing-Remitting (RR) form; RR patients will then evolve into the Secondary Progressive (SP) form after about $10$--$20$ years~\cite{Stamile15a,graphbased}.
Remaining $15$\% of subjects start directly with the Primary Progressive (PP) form.
MS diagnosis has been revolutionized in the last $30$ years by the introduction of MRI.
{\color{\red}Advanced MRI strategies, including magnetization transfer, spectroscopy, and Diffusion Tensor Imaging (DTI), together with classical structural MRI modalities like T1, were successful in detecting alterations and contributed to our understanding of the pathological mechanisms occurring in normal appearing white matter (WM). Indeed, if T1 provides detailed information about the anatomical structure of the gray matter (GM) brain regions, DTI allows to detect the WM connections between those regions.

Determining the current clinical profile of a patient has a major impact on the treatment she gets; unfortunately, it is not an easy task to be carried out with automatic tools.
In the literature, it has been shown that starting from suitable images of the brain, it is possible to extract a structural connectivity matrix, called {\em connectome}, representing the map of neural connections in the brain at hand \cite{DBLP:journals/neuroimage/RubinovS10,graphbased}.
This graph can be expressed as $G=(V,E,\omega)$, where nodes $V$ represent mapped brain regions, edges $E$ represent connections between them and edge weights $\omega$ express the strength of these connections.}
%
%
Recent approaches~\cite{DBLP:conf/icann/Ion-MargineanuK17} exploit state-of-the-art classifiers (such as Support Vector Machines, Linear Discriminant Analysis and Convolutional Neural Networks) to classify Multiple Sclerosis courses using features extracted from Magnetic Resonance Spectroscopic Imaging (MRSI) combined with brain tissue segmentations of grey matter, white matter, and lesions.
Beyond the identification of the current clinical profile, predicting a patient's evolution and response to a therapy based on clinical, biological, and imaging markers still represents a challenge for neurologists.


\subsection{Background on Answer Set Programming}
\label{sec:ASP}
{\color{\red}
The term ``Answer Set Programming'' was introduced by Vladimir Lifschitz to denote a declarative programming methodology~\cite{lifs-99a}; concerning terminology, ASP is sometimes used in a broader sense, referring to any declarative formalism which represents solutions as sets.
However, the more frequent understanding is the one adopted in this article, which dates back to~\cite{gelf-lifs-91}.
For a more thorough introductory material on ASP, we refer the reader to~\cite{Baral:2003:KRR:582493,gelf-leon-02,lifs-99a,mare-trus-99}; in the following, we briefly recall syntax and semantics of the formalism.

The language of ASP is based on rules, allowing (in general) for both disjunction in rule heads and nonmonotonic negation in the body.
It is worth recalling that a significant amount of work has been carried out by the scientific community for extending the basic language, in order to increase the expressive power and improve usability of the formalism.
This has led to a variety of ASP ``dialects'', supported by a corresponding variety of ASP systems\footnote{{\color{\red} During the years, the scientific community has been very active, and many ASP systems have been released relying on different algorithms and solving technologies; we refer the reader to the latest available report on the ASP competition series~\cite{DBLP:journals/corr/abs-1904-09134}, the therein reported references and the vast literature.}} that only share a portion of the basic language; notably, the community relatively recently agreed on the definition of a standard input language for ASP systems, namely ASP-Core-2~\cite{calimeri2012asp}, which is also the official language of the ASP Competition series~\cite{DBLP:journals/aim/CalimeriIKR12,DBLP:conf/aaai/GebserMR16,DBLP:journals/jair/GebserMR17,DBLP:journals/corr/abs-1904-09134}; it features most  of the advanced constructs and mechanisms with a well-defined semantics that have been introduced and implemented in the latest years.


For the sake of simplicity, we next focus on the basic aspects of the language; for a complete reference to the ASP-Core-2 standard, and further details about advanced ASP features, we refer the reader to~\cite{calimeri2012asp} and the vast literature.


A variable or a constant is a {\em term}.
Variables are denoted by strings starting with some uppercase letter, while constants can either be integers, strings starting with some lowercase letter or quoted strings.
An {\em atom} is $a(t_{1}, \dots,$ $t_{n})$, where $a$ is a {\em predicate} of arity $n$ and $t_{1}, \dots, t_{n}$ are terms.
A {\em literal} is either a {\em positive~literal} $p$ or a {\em negative~literal} $\naf p$, where $p$ is an atom.
A {\em disjunctive rule} (or simply {\em rule}, for short) \R{} is a formula of the form:\ \ $ a_1\ \Or \ \cdots\ \Or \ a_n\ $ $ \derives\ b_1,\cdots, b_k,\ $ $ \naf\ b_{k+1},\ \cdots,\ \naf\ b_m. $,\ \ where $a_1,\cdots ,a_n,b_1,\cdots ,b_m$ are atoms and $n\geq 0, m\geq k\geq 0$.
The disjunction $a_1\ \Or\ \cdots\ \Or\ a_n$ is the {\em head} of \R{}, while the conjunction $b_1 , ..., b_k,\ \naf\ b_{k+1} , ...,\ $ $\naf\ b_m$ is the {\em body} of \R{}.
A rule without head literals (i.e.\ $n=0$) is usually referred to as an {\em integrity constraint}.
If the body is empty (i.e.\ $k=m=0$), it is called a {\em fact}.
$H(r)$ denotes the set $\{a_1 ,..., a_n \}$ of the head atoms, and $B(r)$ the set $\{b_1 ,..., b_k, \naf b_{k+1} , \ldots , \naf b_m \}$ of the body literals.
$\BpR$ (resp., $\BnR$) denotes the set of atoms occurring positively (resp., negatively) in $B(r)$.
A rule $r$ is {\em safe} if each variable appearing in $r$ appears also in some positive body literal of $r$.

An {\em ASP program } $\p$ is a finite set of safe rules.
An atom, a literal, a rule, or a program is {\em ground} if no variables appear in it.
According to the database terminology, a predicate occurring only in {\em facts} is referred to as an {\em EDB} predicate, all others as {\em IDB} predicates; the set of facts of $\p$ is denoted by $EDB(\p)$.

The {\em Herbrand Universe} and the {\em Herbrand Base} of $\p$ are defined in the standard way and denoted by $\UP$ and $\BP$, respectively.
Given a rule $r$ occurring in $\p$, a {\em ground instance} of $r$ is a rule obtained from $r$ by replacing every variable $X$ in $r$ by $\sigma (X)$, where $\sigma$ is a substitution mapping the variables occurring in $r$ to constants in $\UP$; $ground( \p)$ denotes the set of all the ground instances of the rules occurring in $\p$.

An {\em interpretation} of $\p$ is a set of ground atoms, that is, an interpretation is a subset $I$ of $\BP$.
A ground positive literal $A$ is {\em true} (resp., {\em false}) w.r.t. $I$ if $A \in I$ (resp., $A \not\in I$).
A ground negative literal $\naf A$ is {\em true} w.r.t. $I$ if $A$ is false w.r.t. $I$; otherwise $\naf A$ is false w.r.t. $I$.
Let $r$ be a ground rule in $ground( \p )$.
The head of $r$ is {\em true} w.r.t. $I$ if $H(r) \cap I \neq \emptyset$.
The body of $r$ is {\em true} w.r.t. $I$ if all body literals of $r$ are true w.r.t. $I$ (i.e., $B^+(r) \subseteq I$ and $B^-(r)\cap I = \emptyset$) and is {\em false} w.r.t. $I$ otherwise.
The rule $r$ is {\em satisfied} (or {\em true}) w.r.t. $I$ if its head is true w.r.t. $I$ or its body is false w.r.t. $I$.
A {\em model} of $\p$ is an interpretation $M$ of $\p$ such that every rule $r \in ground(\p)$ is true w.r.t. $M$.
A model $M$ of $\p$ is {\em minimal} if no model $N$ of $\p$ exists such that $N$ is a proper subset of $M$.
The set of all minimal models of $\p$ is denoted by ${\rm MM}(\p )$.

Given a ground program  $\p$ and an interpretation $I$, the {\em reduct} of $\p$ w.r.t. $I$ is the subset $\p^I$ of $\p$, which is obtained from $\p$ by deleting rules in which a body literal is false w.r.t. $I$.
Note that the above definition of reduct, proposed in~\cite{fabe-etal-2004-jelia}, simplifies the original definition of Gelfond-Lifschitz (GL) transform \cite{gelf-lifs-91}, but is fully equivalent to the GL transform for the definition of answer sets \cite{fabe-etal-2004-jelia}.
Let  $I$ be an interpretation of a program  $\p$.
$I$ is an {\em answer set} (or stable model) of $\p$ if $I \in \mm{\p^I}$ (i.e.,  $I$ is a minimal model of the program $\p^I$)~\cite{przy-91,gelf-lifs-91}.
The set of all answer sets of $\p$ is denoted by  $ANS(\p)$.

\begin{example}
In order to appreciate declarativity and expressiveness of ASP, let us consider the well-known NP-complete $3$-$Coloring$ problem.
Given a graph, we must decide whether there exists an assignment of one out of three colors (red, green, or blue, for instance) to each node such that adjacent nodes always have different colors.
If we suppose that the graph is represented by a set of facts $F$ consisting of instances of the unary predicate $\textit{node}(X)$ and of the binary predicate $\textit{arc}(X,Y)$, then the following ASP program (in
combination with $F$) describes all $3$-$colorings$ (as answer sets) of that graph. 

\smallskip

\begin{footnotesize}
\indent $r_1$:\ \verb" color(X,red) | color(X,green) | color(X,blue) :- node(X). "

\indent $r_2$:\ \verb" :- color(X1, C), color(X2, C), arc(X1, X2). "
\end{footnotesize}

\smallskip

Rule $r_1$ expresses  that each node must either be colored
red, green, or blue\footnote{The same piece of knowledge can be equivalently expressed by means of choice rules, see~\cite{calimeri2012asp}.}; due to minimality of answer sets, a node cannot be assigned more than one color.
The subsequent integrity constraint checks that no pair of adjacent nodes (connected by an arc) is assigned the same color.

Thus, there is a one-to-one correspondence between the
solutions of the 3-Coloring problem and the answer sets of
$F\cup\{r_1, r_2\}$: the graph is 3-colorable if and only if $F\cup\{r_1, r_2\}$ has some answer set.
\end{example}

}

{\color{black}
\section{Framework and Methodology}
\label{sec:framework}

\begin{figure}[t]
\centering
\includegraphics[width=1\textwidth]{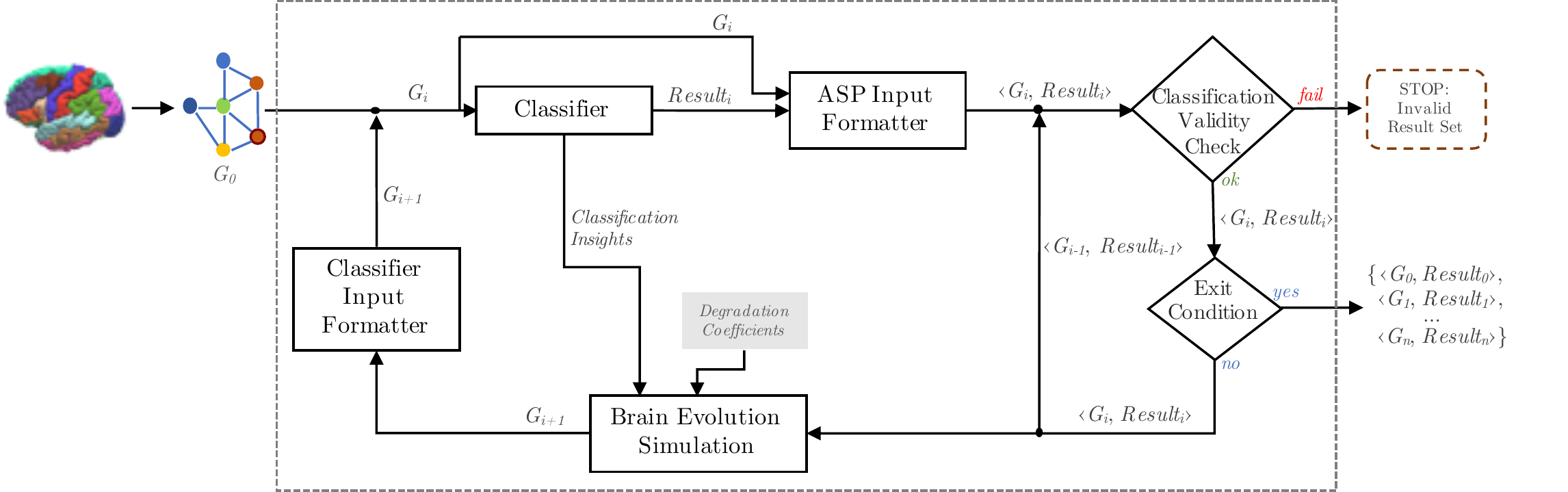}
\caption{Architecture of the proposed framework}
\label{fig:workflow}
\end{figure}

In this section we present a framework to support the analysis of neurological disorders evolution.
{\color{\red}It is worth noting that the framework is intended to be rather general, and can be adapted to several kind of disorders and, as it will be shown in Section~\ref{sec:furtherapplication}, it can be even extrapolated to other scenarios.
Here, we focus on MS disorder.}
We first introduce the general workflow of the framework, and then illustrate its components in abstract terms.
In Section~\ref{sec:specializations} we consider different use cases, and provide some specializations of the various modules.

\smallskip

%
The general workflow is presented in Figure~\ref{fig:workflow}.
Intuitively, it takes a brain representation as input, classifies the current stage of the disease, and then simulates the effects of the disease course by ``damaging the brain''; the newly obtained brain representation is then classified, and used for the next steps.
{\color{\red} These steps are iterated until some condition holds or until a misclassification is detected; in the first case, the overall set of results is provided as output; in the latter case, the execution of the framework is aborted.}

{\color{\red}
More in detail, the framework takes as input a graph representation of the brain of a patient, which can be obtained starting from a combination of MRI acquisition methods.
{\color{\red}As pointed out in Section \ref{sec:background},} this is expected to be a weighted graph $G_{0}=(V,E_0,\omega_0)$ representing the brain connectome of the patient; this graph is then processed by the \emph{Classifier} module.

The {\em Classifier} module can be formally modelled as a function $\chi : \mathcal{G} \rightarrow \mathcal{R}^4$.
In particular, it takes as input the graph $G_{i}$ of the $i$-th iteration of the framework, which represents the possibly modified brain connectome of a patient, and outputs four real values, each indicating the probability of the input graph of expressing a specific MS stage, namely $Result_{i}=(P_{CIS_i}, P_{RR_i}, P_{PP_i}, P_{SP_i})$.
Here, $P_{CIS_i}$ (resp., $P_{RR_i}$, $P_{PP_i}$, $P_{SP_i}$) corresponds to the probability of  $G_{i}$ of representing a CIS (resp., a RR, a PP, a SP) MS stage (see Section \ref{sec:background}).
The stage associated with $G_{i}$ is implicitly the one showing the highest probability.

As previously pointed out, the classification task can be carried out in several ways.
One possibility is to make use of results from graph theory~\cite{DBLP:journals/neuroimage/RubinovS10,graphbased,nlpa}; however, despite network analysis applied on brain connectivity represents a powerful tool, it is not possible yet to define precise biomarkers to classify subjects, especially in the MS context.
As far as current classification methods for MS are concerned, there is a clear distinction between approaches that are based on ANN and those that use different learning methods, such as Support Vector Machines (SVM).
%
ANN demonstrated to be one of the most promising tools for the analysis and classification of images, and has been used in a wide range of applications even if, to date, image analysis via ANN in the context of MS has been exploited mostly for the identification of MS lesions rather than MS profile classification.
%
%
As usually done in similar contexts, in the approaches studying MS via ANN the ground truth for training purposes is obtained by annotations on the data directly provided by human experts. The ANN training phase is a pre-processing step propaedeutic to the application of the framework.

Depending on the actual implementation of the classifier, this module may also provide some insights on the classification process that could be of use in the subsequent steps.
This case will be detailed in Section~\ref{sec:specializations}.
Intuitively, the specific ANN we adopt in the specialization of our framework allows to leverage the topology of the graph for the classification process and to compute a meaningful coefficient for each graph edge, in the form $(x,y,imp_{xy})$, representing the importance of the edge $(x,y)$ in the classification task.
These insights can be exploited in the simulation of brain evolution and are expressed in Figure~\ref{fig:workflow} by the direct connection between the \emph{Classifier} module and the \emph{Brain Evolution Simulation} module.

\smallskip


The output of the classification is verified by a {\em Classification Validity Checker}.
{\color{\red} One of the recently arising research issues in  classification tasks is the automatic check of validity of results.
Indeed, {\color{\red} verification of ANN results is receiving increasing interest, and it is currently a very active research area~\cite{PuTa*10,DBLP:journals/corr/abs-1811-11373,DBLP:journals/corr/abs-1805-09938}.}}
{\color{\red} However, given the nature of the problem we are addressing in this paper, we can simplify this task by resorting to rules and constraints that model domain knowledge.
}

In particular, if certain clinical evidence is available for a patient, this can be used to limit classification alternatives.
As an example, it is well known that a brain model obtained by a severe disruption of its previous structure cannot induce a remission of the pathology.
In other words, given a brain structure $G_{i-1}$ classified, e.g., as RR, and given its modified version $G_{i}$ obtained by a strong reduction of arcs and/or weights, if $G_{i}$ is classified as CIS, i.e., a remission is hypothesized from $G_{i-1}$ to $G_{i}$, the classification is evidently wrong and must be discarded. {\color{\red}This kind of checks} can be easily carried out through suitable sets of rules and constraints; these will be presented in detail in Section~\ref{sec:specializations}.
As a side note, if the starting input graph $G_0$ is annotated with the ground truth provided by an expert on the right classification of the initial MS stage, this step can stop the process at the very first iteration if the classification result disagrees with the ground truth.
In order to provide the input to the checker in the proper formalism, an {\em ASP Input Formatter} module stands between the classifier and the checker.
It is in charge of translating the output provided by the classifier and the current graph into ASP facts.

The {\em Classification Validity Checker} can be modelled in abstract terms as a function  $\nu : \mathcal{G} \times \mathcal{R}^4 \times \mathcal{G} \times \mathcal{R}^4 \rightarrow \{$``OK''$, $``FAIL''$ \}$ which takes as input two graphs, $G_{i-1}$ and $G_{i}$ and the corresponding classification results $Result_{i}=(P_{CIS_i}, P_{RR_i}, P_{PP_i}, P_{SP_i})$ and $Result_{i-1}=(P_{CIS_{i-1}}, P_{RR_{i-1}},$ $P_{PP_{i-1}}, P_{SP_{i-1}})$; it provides as output one among the two possible values ``OK'' or ``FAIL'', depending on the outcome of the check.
%
%
If the Classification Validity Checker returns ``FAIL'', the iterative process is immediately stopped and current and previous results are invalidated. On the contrary, if the checker returns ``OK'', the execution of the framework proceeds to the next steps.
	
In particular, a generic {\em Exit Condition} is subsequently checked in order to verify whether it is necessary to proceed with the next iteration of the framework or not.
The definition of such condition strongly depends on the objective of the analysis.
As an example, it could be interesting to check if a certain target probability is reached for a certain MS stage, or if a certain degree of disruption of the original graph has been induced in the last step, or simply if the required number of iterations has been carried out.
If the exit condition is verified, the execution is stopped and the set $\{\langle G_0, Result_0 \rangle, \langle G_1, Result_1 \rangle, \ldots \langle G_n, Result_n \rangle \}$ of graphs and corresponding classification results are provided as output. Otherwise, the execution proceeds with the next brain evolution simulation step.

The {\em Brain Evolution Simulation} module can be formally modelled as a function $\mu: \mathcal{G} \times 2^{E} \times 2^{E} \rightarrow \mathcal{G}$; here, $\mathcal{G}$ is the set of all possible graphs, whereas $2^{E}$ represents all possible sets of triples of the form $(x,y,c_{xy})$, where $x$ and $y$ are nodes of the graph and $c_{xy}$ is a label.
In particular, $\mu$ takes as input the current graph $G_{i}$, and two sets of triples $(x,y,imp_{xy})$, and $(x,y,dc_{xy})$ which convey, respectively, information on the importance of each edge for the classification task, whenever provided by the classifier, and information on how to modify edge weights during the simulation.
We formally introduce and specialize these sets is Section~\ref{sec:specializations}.
$\mu$ provides as output a new graph $G_{i+1}$ which represents a simulated evolution of brain structure.

Some metrics over graphs representing brain structures have been considered in previous studies on MS~\cite{DBLP:journals/neuroimage/RubinovS10}; however, it is still unclear how these metrics influence the progress of the disease.
In our framework, the {\em Brain Evolution Simulation} module is used to explore a wide variety of graph modification criteria.
In this context, ASP plays a very relevant role as a fast and effective tool for the definition and identification of subgraphs satisfying some predefined property that could be involved in MS course; in some cases, the identification of such subgraphs may involve the solution of optimization problems, whose coding can be significantly time-consuming in other programming paradigms. In our framework, the \emph{Brain Evolution Simulation} module   is composed of an ASP program of choice that, given the graph $G_{i}$, first defines a connectome modification criterion described by a set of edges to modify and then produces the new graph $G_{i+1}$. The corresponding ASP program(s) enjoy the nice properties of such a declarative formalism, resulting very flexible and easy to adapt to small changes in the desiderata.

Each ASP program  is coupled with an extensional knowledge base consisting of a set of facts  representing nodes and edges of $G_i$, and identifies a set of atoms  which represent the set of edges $E'$ to be modified.
Given that, in our context, edge weights are related to the number of fibers linking two points in the brain, an edge $(x,y,w)$ with $w=0$ is considered inactive and not contributing to the network, i.e., the corresponding nodes are considered not connected.
If available, the choice of $E'$ can also be guided by the information about the importance of each edge for the classification task; this information is expressed by the first set of triples $(x,y,imp_{xy})$ provided as input along $G_{i}$.

As already noted, different ways of altering the brain structure using $E'$ can be devised.
A basic altering method could consist in simply removing these arcs; this corresponds to set the weight of each edge in $E'$ to 0.
However, edge weights play an important role for the structure itself and, from a biological point of view, the strength of the connections (expressed by edge weights in our model) progressively decreases while the brain degenerates.
As a consequence, possible evolution strategies might include progressive variations of selected edge weights, as expressed by the second set of triples $(x,y,dc_{xy})$ provided as input. All the edges not included in $E'$ are simply copied into $G_{i+1}$.

Examples of interesting criteria for identifying the set of edges to be modified, and that will be detailed in Section~\ref{sec:specializations}, are reported next:
\begin{compactitem}
\item[$(i)$]
    \emph{Max Clique}: contains the greatest subset of vertices in $G$ such that every two distinct vertices in the clique are connected by an edge;
\item[$(ii)$]
    \emph{$k$-hub}: the set of $k$ nodes having the highest degree;
\item[$(iii)$]
    \emph{Min Vertex Cover}: the smallest set of vertices $MVC$ such that each edge of the graph is incident onto at least one vertex of the set;
\item[$(iv)$]
    \emph{Density reduction}: the minimal set of edges that, if removed, allows a reduction of the graph density by a given amount;
\item[$(v)$]
    \emph{Assortativity increase}: the minimal set of edges that, if removed, allows an increase of the graph assortativity by a given amount.
\end{compactitem}

It is important to note that, as already mentioned and as it will be clearer in the following, switching between these properties or slightly modifying the criteria in ASP requires just to change a few rules; on the contrary, using a classical imperative programming scheme, it would require to rewrite and adapt source code that can be significantly harder to maintain.

\smallskip

The newly obtained graph $G_{i+1}$ is given as input back to the \emph{Classifier}; the ASP representation of $G_{i+1}$ is translated back into the format required by the \emph{Classifier} by the {\em Classifier Input Formatter} module.

}

\section{Specializations of the framework}
\label{sec:specializations}

{\color{\red} In this section} we specialize the framework to three biologically relevant settings, {\color{\red} upon which we also carry out some experiments in Section~\ref{sec:experiments}.}
Some of the modules are implemented in the same way for all the three specializations, and are hence described first.

Before the actual description, it is worth pointing out, once again, that the goal of the present work is not to provide clinical validation of some kind of results; rather, we want to show the potential of the herein proposed framework that, thanks to the {\color{\red} combined use of} ASP and ANN, can help experts in studying the evolution of the disease from different perspectives.


\subsection{From MRI to Graphs}
\label{sub:mri-graphs}

This is a pre-processing step needed to transform brain images into graphs.
Starting from a combination of MRI acquisition methods like T1 and DTI {\color{\red}(see Section \ref{sec:neurological})}, it is possible to extract structural connectivity matrices, representing topological features of the brain \cite{DBLP:journals/neuroimage/RubinovS10,graphbased}.
The graph generation pipeline used to extract a brain network can be divided in three main steps:

\begin{enumerate}
	\item
	Parcellation of the cortical and sub-cortical grey matter (GM) is performed on 3D T1-weighted images in order to label T1 voxels in four groups [white matter (WM), cortical GM, sub-cortical GM, cerebro-spinal fluid (CSF)] and then to define graph nodes.
	\item
	Diffusion weighted MRI images are preprocessed using correction of Eddy-current distortions \cite{DBLP:journals/neuroimage/JenkinsonBBWS12} and skull stripping. Probabilistic streamline tractography \cite{DBLP:journals/imst/TournierCC12} on diffusion images is then applied to generate fiber-tracks in voxels labeled as WM voxels.
	\item
	{\color{\red} The connectivity matrix $A \in \mathbb{R}^{q\times q}$ $(q = 84)$ is generated for each subject. More formally, $A$ represents the adjacency matrix of the weighted undirected graph $G=(V,E,\omega)$ where $V$ is the set containing the segmented GM brain regions (with $|V| = q$), $E$ is the set of graph edges defined as: $$E = \{\{i,j\}~|~\omega(i,j) > 0,~1 \le i,j \le q\}$$ and $\omega : \mathbb{N}^{2} \to [0,1]$ is a function that measures the strength of the connection between a pair of nodes by summing the number of streamlines connecting them and scaling this number in the range $[0,1]$.\footnote{It is worth observing that, since current ASP systems do not support real numbers, the ASP Input Formatter scales values in the real interval $[0,1]$ into integer values between $0$ and $100$.}}

\end{enumerate}

Figure \ref{fig:graphconstruct} illustrates the whole process.

\begin{figure}[t!]
\centering
    \includegraphics[width=0.85\textwidth]{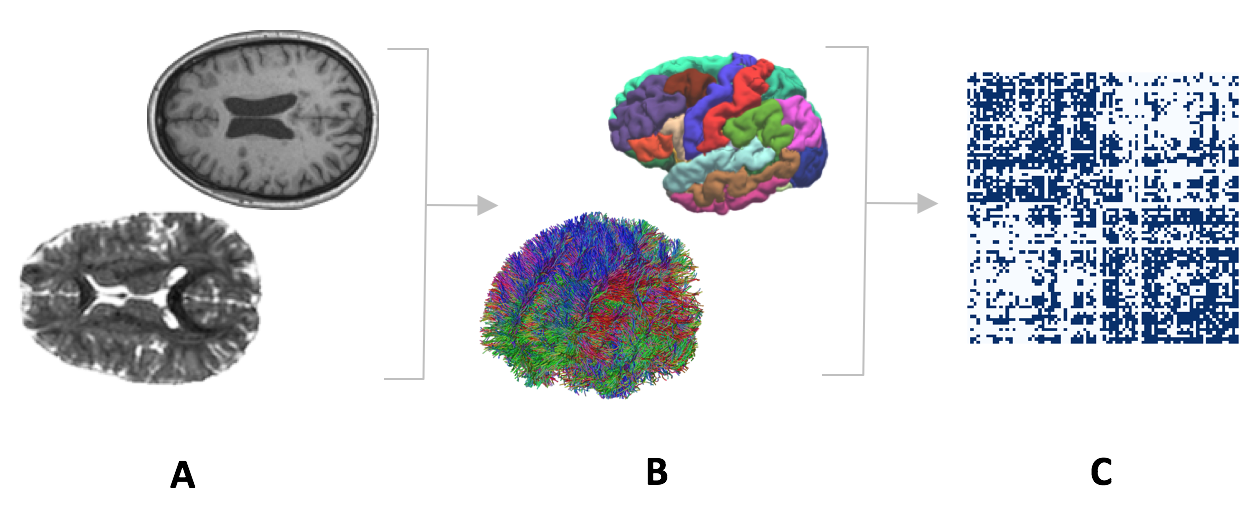}
    \caption{Illustration of the graph creation steps: (A) T1 and diffusion weighted MR images are used to generate cortical parcellation and fiber tractography (B), which are combined to generate connectivity matrix (C)\label{fig:graphconstruct}}	
\end{figure}

\subsection{Specialization of the Classifier}
\label{sub:NN-spec}
The Classification of MS patients in their respective clinical forms is achieved by means of a slightly modified version of BrainNetCNN~\cite{DBLP:journals/neuroimage/KawaharaBMBCGZH17} we specifically defined for this work.
BrainNetCNN is a convolutional neural network (CNN) framework which operates on brain connectivity.
Differently from the spatially local convolutions done by traditional CNN, BrainNetCNN is designed to exploit topological locality of structural brain networks.
This result is achieved by using two novel operators: the \textit{Edge2Edge} layer, which performs a convolution over the weights of edges that share nodes together, and the \textit{Edge2Node} layer, which computes, for each node $i$, a weighted combination of the incoming and outgoing weights of edges connected to $i$.

{\color{\red}The BrainNetCNN architecture parameters originally proposed by Kawahara \emph{et al.} were kept unchanged since they already showed promising results in predicting clinical neurodevelopmental outcomes from brain networks.} The model takes as input the adjacency matrix representation of a graph $G$ (the brain connectivity matrix) and outputs a probability value for each MS stage. The architecture is composed of two \textit{Edge2Edge} layers, which process the input using 32 filters, followed by two \textit{Edge2Node} layers with 64 and 256 filters.
Then, two \textit{fully connected} layers of size 128 and 30 are applied.
Both convolutional and fully connected layers use \emph{Leaky ReLU} ($\alpha$ = 0.33) activation function.
Finally, a fully connected layer of size 4 (the output layer) with \emph{softmax} activation is used to perform the classification.

In Section~\ref{sec:training-classification} we show that this new version of the classifier considerably increases reliability with respect to previous proposals ~\cite{DBLP:conf/esann/CalimeriMST18}.
This is important in our framework, as the precision of the classification over unseen samples is fundamental in studying MS evolution over brain alterations.
Moreover, as it will be specified in Section \ref{sub:insights-spec}, this ANN allows to compute a coefficient for each graph edge that represents its importance in the classification task.

\subsection{{\color{\red}Specialization of the Classification Validity Checker}}
\label{sub:checker-spec}

{\color{\red} The Classification Validity Checker is implemented via an ASP encoding.
Due to pace reasons, we refrain from discussing in detail this and the following encodings, but we point out that they comply with the ASP-Core-2 standard; the interested reader can refer to Section \ref{sec:ASP} and the vast literature on Answer Set Programming for more details.}

{\color{\red} For the purpose of this work we implement a very basic checker that can be easily enriched with more domain specific knowledge; this is shown in Figure \ref{alg:checker}.
In particular, it takes as input two graphs $G_{i}$ and $G_{i-1}$ encoded by a set of facts of the form \texttt{edge(X,Y,W)} and \texttt{edge\_1(X,Y,W)}, respectively; it takes as input also a threshold $T$, used to determine whether a severe disruption occurred and the results obtained by the classifier at steps $i$ and $i-1$, encoded as facts of the form {\tt result(STAGE, P)} and {\tt result\_1(STAGE, P)}, where {\tt STAGE} $\in \{$``CIS'', ``RR'',``PP'', ``SP''$\}$ and {\tt P} represents the probability computed for the corresponding stage by the classifier.}

{\color{\red} The encoding first determines whether a severe disruption occurred in the brain model between $G_{i-1}$ and $G_{i}$ by counting the number of arcs that have been removed from $G_{i-1}$ and comparing it with a threshold.
If this happens, it checks whether a known impossible transition has been inferred by the classifier between step $i-1$ and step $i$.
As an example, in presence of a severe disruption, it is well known that a transition from RR (resp., from PP, or SP) to CIS is biologically implausible \cite{lublin2014defining}. If a known impossible transition is detected, the computed answer set contains \texttt{check("FAIL")}; otherwise, the answer set contains \texttt{check("OK")}.}

\begin{figure}[t]
\begin{footnotesize}
{\color{\red}
	\begin{verbatim}
	% input: facts of the form "edge(X,Y,W)" and "edge_1(X,Y,W)" as the input graphs
	% input: facts of the form "result(STAGE,P)",
	%    e.g., result("CIS", 90). result("RR", 25). result("PP", 15). result("SP", 10).
	%        representing the classification results for the current graph
	% input: facts of the form "result_1(STAGE,P)",
	%        representing the classificatoin results for the previous graph
	% input: a fact of the form "Th(T)" indicating the minimum number of removed arcs
	%        representing a severe disruption according to domain knowledge

	% determine if a severe disruption occurred in the last iteration
	severedisruption :- #count{X,Y: edge_1(X,Y,W1), edge(X,Y,W), W1>0, W=0}>T, Th(T).
	
	% check for validity of the classification step
	check("FAIL") :- severedisruption, result("CIS",R_CIS), result_1(S,R_1), S!="CIS",
	                 #max{ R : result(_,R)} = R_CIS,  #max{ R : result_1(_,R)} = R_1.

	check("OK") :- not check("FAIL").
	\end{verbatim}
	\caption{An ASP encoding for the Classification Validity Checker.\label{alg:checker}}
}
\end{footnotesize}
\end{figure}


%
%
%
%

\subsection{Specialization of the Exit Condition}
\label{sub:exit-spec}

{\color{\red} The exit condition can be specialized in several ways. 
As an example, it can stop iterations as soon as it detects that the classifier predicts a transition from one MS stage to another for the current patient, for instance when a patient starts from a CIS stage and, after some modifications to the brain structure, she is classified as RR.
This would imply that modifications simulated to the brain structure are sufficient to simulate a transition in the pathology of the patient. 
Analogously, the exit condition can stop the iterations of the framework whenever the difference between predicted probabilities become very low (i.e., below a certain threshold) from one iteration to the other; this may imply that the last modifications simulated on the brain structure are no more informative.

In our framework, we exploited a simple exit condition which stops the execution after a certain number of iterations.}

\subsection{Specializations for three different use cases}

\subsubsection{Specialization for studying structural properties}
\label{sub:structural-spec}


A first interesting use case for our framework is the study of the impact of graph structural properties, and their modification, in the evolution of MS. 
The aim is to determine whether there is a latent relation between the presence/absence of particular graph structures in the connectome and the stage of the MS disease.
In particular, we are interested in verifying if and how modifications on the connectome of a patient, simulated by modifications on the graph representing it, can modify the classification returned by the ANN. 
Observe that, as a side effect, understanding these relations could provide at least partial motivations for ANN classifications; this is still an open issue in ANN.

{\color{\red}In this use case, the Brain Evolution Simulation module must be specialized to detect the structural property of interest, and in particular the set of edges representing it, and to generate a new connectome by modifying selected edges.}

{\color{\red} Interesting criteria and returned edges are the following: 
\emph{(i)} \emph{Max Clique}, i.e., the greatest subset of vertices in $G$ such that every two distinct vertices in the set are adjacent; in this case, the module modifies the edges $E'$ linking the vertices in the clique. 
\emph{(ii)} \emph{Independent Set}, i.e., the greatest subset of vertices in $G$ such that no two vertices in the set are adjacent; in this case the module modifies the edges $E'$ having exactly one vertex in the independent set. 
\emph{(iii)} \emph{Max-degree node}, i.e., the node showing the maximum degree in $G$; in this case the module modifies the edges connected to it. 
\emph{(iv)} \emph{$k$-hub}, i.e., the set of $k$ nodes having the highest degree; in this case the module modifies the edges connected to the \emph{$k$-hub}.
\emph{(v)} \emph{Min Vertex Cover}, i.e., the smallest set of vertices $MVC$ such that each edge of the graph is incident onto at least one vertex of the set; in this case the module modifies the edges $E'$ such that both vertices of the edge are in $MVC$.}

Each version of the Brain Evolution Simulation module is then obtained by a suitable ASP encoding detecting the property and the corresponding edges. As an example, Figure~\ref{alg:clique} reports an ASP encoding for the \emph{Max Clique} problem. 
%

{\color{\red}
Intuitively, the program ``guesses'' the nodes that belong to a clique in the graph $G_i$ by means of the choice rule:

\texttt{\{clique(X)\} :- node(X)}

\noindent and then checks, by means of the strong constraint:

\texttt{:- clique(X), clique(Y), X < Y, not activeEdge(X,Y)}

\noindent that the inclusion of two unconnected nodes in the candidate clique set is forbidden.
Cardinality of the clique is maximized using the weak constraint

\texttt{:\~\ node(X), not clique(X). [1@1,X]}

\noindent that penalizes the exclusion of a node in the candidate clique set.
}

{\color{\red}The set of the edges connecting the nodes within the resulting clique is represented by the extension of predicate \texttt{e(X,Y,W)}, which is built according to the rule \texttt{e(X,Y,W) :- edge(X,Y,W), clique(X), clique(Y)}.
The new modified graph $G_{i+1}$ is built with the last two rules appearing in the encoding. 
In particular, all the edges that must not be modified are just copied in the new graph $G_{i+1}$ (see the last but one rule).
The last rule simulates the progressive disrupting process of the MS disease on the portion of brain connectome identified by the extension of predicate \texttt{e(X,Y,W)}; specifically, we designed it in order to act as a degradation function on the weights of selected edges; this simulates a degradation in the strength of the connections.
In particular, given the initial graph $G_0$ a {\em degradation coefficient} is computed for each edge $(x,y,w_{xy})$ in $G_0$ as $d_{xy}=w_{xy}\times p$, where $p$ is a percentage of degradation, set as a parameter for the experimentation.
Degradation coefficients are given as input to the program as facts of the form \texttt{dc(X,Y,D)} and computed as a preprocessing step before starting framework iterations. 
Then, each edge \texttt{e(X,Y,Wxy)} generates  in  $G_{i+1}$ an edge \texttt{edge1(X,Y,max\{Wxy-Dxy,0\})} (see the last rule).
Here, a weight set to $0$ means a deletion of the edge from the resulting graph (it will no longer be an {\tt activeEdge}), and consequently a complete disruption of the corresponding connection; in this case, the subsequent iterations and the corresponding ASP programs will no longer consider this edge as belonging to the graph.
In our experiments, we considered both $p=25\%$ and $p=50\%$.

The value of $p$ heuristically determines the intensity of degradation applied to the strength of the connections in one iteration and, consequently, the velocity of degradation of the connectome through the iterations of the framework. 
Thus, the choice of $p$ determines the velocity of the simulation. 
In particular, if we assume that $p=25\%$ and that the same edge is chosen at each iteration\footnote{Observe that this is not obvious, and it strongly depends on both the studied property  and the configurations of the other edges.}, it takes four iterations to virtually remove it from the connectome.
Analogously, when $p=50\%$, and the same edge is chosen at each iteration, two iterations of the framework are sufficient to virtually remove it. 
As a consequence, the choice of $p$ strictly depends on the granularity of degradations one wants to study and on the maximum number of iterations of the framework that one wants to carry out.
}

All ASP encodings for considered criteria can be found at~\additionalmaterial.

\begin{figure}[t]
\begin{footnotesize}
    {\color{\red}
    \begin{verbatim}
    % input: facts of the form "node(X)" and "edge(X,Y,W)" as the input graph
    % input: facts of the form "dc(X,Y,D)" as degradation coefficients for all edges
    % input: support atom "zero(0)"

    % guess the clique
    { clique(X) } :- node(X).

    % take into account only active edges
    activeEdge(X,Y) :- edge(X,Y,W), W>0.
    :- clique(X), clique(Y), X < Y, not activeEdge(X,Y).

    % maximize clique cardinality
    :~ node(X), not clique(X). [1@1,X]

    % edges to be modifed in the new graph
    e(X,Y,W) :- edge(X,Y,W), clique(X), clique(Y).

    % define the new graph
    edge1(X,Y,W) :- edge(X,Y,W), not e(X,Y,W).
    edge1(X,Y,NW) :- e(X,Y,W),
                     #max{K : e(X,Y,W), dc(X,Y,D), K=W-D ; 0 : zero(0)} = NW.
    \end{verbatim}
    }
    \caption{An ASP encoding for the \emph{Max Clique} problem.\label{alg:clique}}
\end{footnotesize}
\end{figure}

\subsubsection{Specialization for studying graph metrics}
\label{sub:metrics-spec}

\begin{figure}[t]
	\centering
	\includegraphics[width=.7\textwidth]{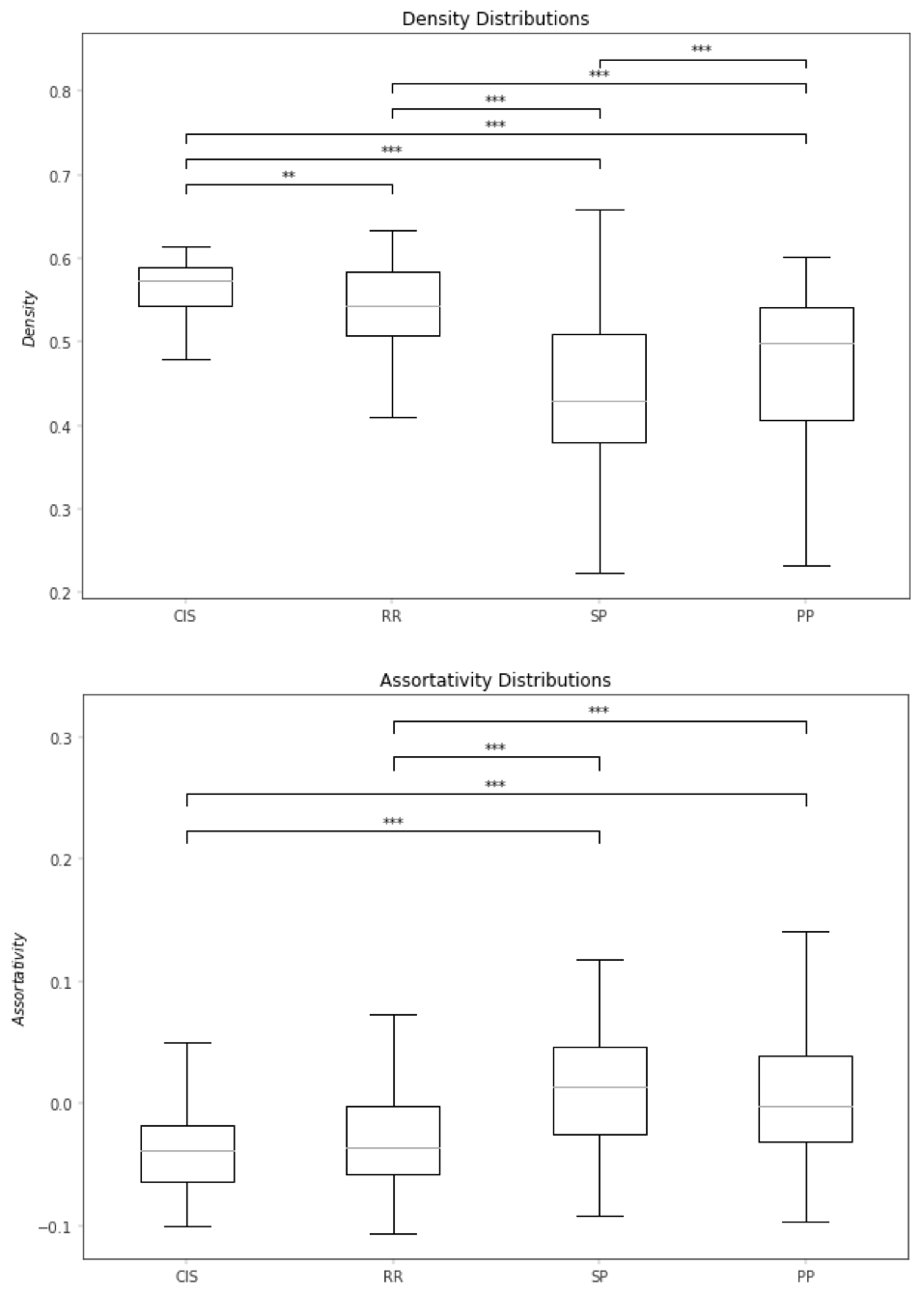}
	\caption{Summary of results on the correlation between Density and Assortativity and MS stages.}
	\label{fig:metrics-dependency}
\end{figure}

In~\cite{graphbased} a relationship between some graph metrics, e.g., Density and Assortativity, and the MS stage of a patient has been clinically demonstrated (see Figure~\ref{fig:metrics-dependency} for a summary). 
It is then interesting to evaluate whether the evolution of the disease could be also related to progressive modifications (decrease/increase) of such metrics obtained by modifications in the corresponding graphs. 
It is worth pointing out that a change in these metrics is not always related to specific substructures {\color{\red}and, generally, depends on presence/absence of edges.}
As a consequence, the setting introduced in the previous section cannot be applied in this new context.
Moreover, while searching for a variation in a metric, as an example a decrease in density, it is important to avoid trivial modifications such as the removal of all the edges. 
It is also worth noting that, for some metrics such as assortativity, the removal of edges may induce either a decrease or an increase in the property.
As a consequence, it is crucial to look at minimal graph changes allowing to reach a certain goal in metric variation.
{\color{\red} Also in this context, the expressiveness and compactness of ASP and its extensions allow for very elegant and readable encodings even for optimization problems, i.e., the identification of minimal changes involving the computation of complex metrics.}

{\color{\red} Among the metrics considered in~\cite{graphbased}, we focus on Density and Assortativity. In order to keep the paper self contained, we next recall the basic definition of these metrics.

The density $d$ of a graph $G=\langle V, E \rangle$ is defined as:

$$ d=\frac{|E|}{|V|(|V|-1)} $$

\noindent where $|E|$ is the number of edges in $G$ and $|V|$ represents the number of nodes in $G$.

Assortativity measures the similarity of connections in the graph with respect to the node degree. 
In particular, the formula for computing the assortativity degree of a graph $G$ is defined as \cite{Newman2002}:
$$ r=\frac{\sum_{xy}(xy(e_{xy}-a_x  b_y))}{\sigma_{a}\sigma_{b}} $$ 

\smallskip

\noindent where $x$ and $y$ are values of node degrees for $G$,\ $e_{xy}$ is the fraction of all edges in $G$ joining vertices having degree values $x$ and $y$,\ and $a_{x}= \sum_{y}e_{xy}$,\ $b_{y}=\sum_{x}e_{xy}$. Moreover, $\sigma_{a}$ and $\sigma_{b}$ are the standard deviations of the distributions $a_x$ and $b_y$.

The formulas above, especially the one for assortativity, show that the computation of graph metrics can be in general not easy to be carried out using only rules in an ASP program.}
{\color{\red}As a consequence, we resort to recent extensions of ASP systems, such as I-DLV/DLV2~\cite{DBLP:journals/ia/CalimeriFPZ17,Adrian2018-KI-dlv2}, DLVHEX~\cite{DBLP:journals/tplp/EiterFIKRS16,DBLP:journals/tciaig/CalimeriFGHIR0T16,eggkrsw2018-ki}, etc., which allow the integration of external computation sources within the ASP program. 
In particular, the problem at hand requires to send a (possibly guessed) entire graph, i.e., a set of edges, to an external source of computation.
The ASP standardization group has not released standard language directives yet for such features; here, we make use of syntax and semantics of DLVHEX~\cite{DBLP:journals/tplp/EiterFIKRS16}, while a slightly different formulation must be used to comply with I-DLV/DLV2~\cite{DBLP:journals/ia/CalimeriFPZ17,Adrian2018-KI-dlv2} or clingo~\cite{DBLP:journals/tplp/GebserKKS19} syntax.
}

{\color{\red}
Figure~\ref{alg:decrease} shows the specialization of the Brain Evolution Simulation module for deriving the minimal changes to perform on a graph $G_{i}$ in order to obtain a decrease in the measure of a certain property (by 10\% by edge removal in the example).

In particular, the program of Figure~\ref{alg:decrease} first defines the current value of the metric on the input graph; this is done with the support of a call (via ``external'' atom \verb+&computeMetric+) to an external function written in an imperative programming language, e.g., Python.
Then, it defines the target value for the metric.
Hence, the program reports the guess for a subgraph that satisfies the goal, where the number of removed edges, expressed by the edges not \verb+in+ the subgraph, is minimized by the weak constraint.
Finally, the new graph $G_{i+1}$ is generated by the last two rules. Specifically, edges of $G_{i}$ in the guessed subgraph are copied into $G_{i+1}$, whereas all the other edges are removed by setting their weight to 0.
}
The changes to be performed on the program for analyzing an increase of the metric are straightforward.

\begin{figure}[t]
\begin{footnotesize}
{\color{\red}
	\begin{verbatim}
	% input: facts of the form "node(X)" and "edge(X,Y,W)" as the input graph
	% compute the starting value of the metric and the corresponding threshold
	target(X) :- &computeMetric[activeEdge](Y), X=Y/90.
	
	% take into account only active edges	
	activeEdge(X,Y) :- edge(X,Y,W), W>0.
	
	% guess the new graph
	0 < { in(X,Y): activeEdge(X,Y) }.

	% check that the goal is reached
	:- &computeMetric[in](X), target(K), X > K.
	
	% minimize removed edges
	:~ activeEdge(X,Y), not in(X,Y). [1@1,X,Y]

	% define the new graph
	edge1(X,Y,W) :- in(X,Y), edge(X,Y,W).
	edge1(X,Y,0) :- edge(X,Y,W), not in(X,Y).

	\end{verbatim}
	\caption{An ASP encoding for analyzing the decrease of a graph metric.\label{alg:decrease}}
}	
\end{footnotesize}
\end{figure}



\subsubsection{Specialization for exploiting ANN insights}
\label{sub:insights-spec}
As previously pointed out, the specific ANN presented in Section \ref{sub:NN-spec}, leverages the topology of the graph for the classification process. Furthermore, the particular structure of the network allows to estimate a meaningful coefficient for each graph edge representing the importance of that edge in the classification task. Specifically, in order to define an importance coefficient for each input edge of the brain connectivity,
 we used the method formerly proposed by Simonyan \emph{et al.}~\cite{Simonyan} and already applied to BrainNetCNN by Kawahara \emph{et al.}~\cite{DBLP:journals/neuroimage/KawaharaBMBCGZH17}.

Let $p_{c}(G)$ be the score assigned to the class $c$ by the trained classification layer of the ANN for a given graph $G=(V,E,\omega)$. Then, the contribution of each edge $e \in E$ can be estimated based on its influence to the score $p_{c}(G)$. More in detail, the edge importance is computed by the partial derivative $\frac{\delta{p_{c}(G)}}{\delta{e}}$ for each edge $e \in E$ by backpropagation~\cite{Simonyan}.


{\color{\red} Then, given a graph $G$, let us assume we can derive from the output of the Classifier module a set of facts of the form \texttt{imp(X,Y,P)}, where $P$ is the importance in the ANN of the edge $(X,Y)$ given by the formula introduced above.
It is strongly interesting to study the impact over MS clinical course of both structural properties over graphs and graph metrics also taking into account these insights from the ANN.
Indeed, determining that there are sets of edges that could be ignored without losing important information or, conversely, that it is possible to focus on a small subset of edges, could both significantly help experts in their analyses and reduce computational requirements.}

{\color{\red} The specialization of the framework taking into account edge importance is quite straightforward.
In particular, let us use a threshold $T$ to distinguish between important ($P \geq T$) and unimportant ($P \leq T$) edges.
Consider the ASP program shown in Figure~\ref{alg:clique-imp}; this is substantially the same as the one shown in Figure~\ref{alg:clique} for studying cliques, except for the second rule, which forces important edges (if $P\geq T$ is used) or unimportant edges (if $P\leq T$ is used) to belong to the clique.
The threshold $T$ can be dynamically set.
Observe that putting unimportant edges in the clique, means that the graph will be modified on unimportant parts only (as judged by the ANN).
Differently from what we have shown in Figure~\ref{alg:clique}, the last two rules state that, in the new graph, the weights of edges identified by the clique are set to $0$ and, hence removed.}

\begin{figure}[t]
\begin{footnotesize}
{\color{\red}	
	\begin{verbatim}
	% input: facts of the form "node(X)" and "edge(X,Y,W)" as the input graph
	% input: facts of the form "imp(X,Y,P)" denoting the importance of edge(X,Y,W)
	% input: a fact of the form Th(T) used to discriminate between important and
	%       unimportant edges
	% guess the clique
	{ clique(X) } :- node(X).

	% take into account only active edges
	% if P <= T unimportant edges are in the clique,
	% if P >= T important edges are in the clique
	activeEdge(X,Y) :- edge(X,Y,W), W>0, imp(X,Y,P), Th(T), P <= T.
	:- clique(X), clique(Y), X < Y, not activeEdge(X,Y).
	
	% maximize clique cardinality
	:~ node(X), not clique(X). [1@1,X]
	
	% edges to be modifed in the new graph
	e(X,Y,W) :- edge(X,Y,W), clique(X), clique(Y).
	
	% define the new graph
	edge1(X,Y,W) :- edge(X,Y,W), not e(X,Y,W).
	edge1(X,Y,0) :- e(X,Y,W).
	\end{verbatim}
	\caption{An ASP encoding for the \emph{Max Clique} problem maximizing the use of important/unimportant edges.\label{alg:clique-imp}}
}
\end{footnotesize}
\end{figure}

Analogously, Figure \ref{alg:decrease-imp} shows the specialization for graph metrics by minimizing/maximizing the use of important edges in reaching the goal for graph metrics variation. Again, there is a minimal difference consisting in the introduction of a weak constraint that minimizes the use of important/unimportant edges.

\begin{figure}[t]
\begin{footnotesize}
{\color{\red}	
	\begin{verbatim}
	% input: facts of the form "node(X)" and "edge(X,Y,W)" as the input graph
	% input: facts of the form "imp(X,Y,P)" denoting the importance of edge(X,Y,W)
	% input: a fact of the form Th(T) used to discriminate between important and
	%        unimportant edges
	% compute the starting value of the metric and the corresponding threshold
	target(X):- &computeMetric[activeEdge](Y), X=Y/90.
	
	% take into account only active edges
	activeEdge(X,Y):- edge(X,Y,W), W>0.

	% guess the new graph
	0 < { in(X,Y): activeEdge(X,Y)}.

	% check that the goal is reached
	:- &computeMetric[in](X), target(K), X > K.
	:~ activeEdge(X,Y), not in(X,Y). [1@2,X,Y]

	% In the next constraint,
	% if P >= T maximizes removal of unimportant edges,
	% if P <= T maximizes removal of important edges
	:~ activeEdge(X,Y), not in(X,Y), imp(X,Y,P), Th(T), P >= T. [1@1,X,Y]

	% define the new graph
	edge1(X,Y,W) :- in(X,Y), edge(X,Y,W).
	edge1(X,Y,0) :- edge(X,Y,W), not in(X,Y).
	\end{verbatim}
	\caption{An ASP encoding for analyzing the decrease of a graph metric maximizing the use of important/unimportant edges.\label{alg:decrease-imp}}
}
\end{footnotesize}
\end{figure}



\section{Experiments}
\label{sec:experiments}

{\color{\red}This section reports about the experiments we carried out in order to assess the proposed framework and its specializations.
The section is organized in three parts.
The aim of the first part is to show the flexibility of the framework, and to adapt the analysis to different perspectives.
In particular, we focus on analyzing how classification probabilities vary during the brain evolution simulation, in order to verify appropriateness of the approach in the biomedical context herein discussed.
The second part specifically focuses on performance, while the third part is devoted to the discussion of  obtained results and to outline  some lessons learned thanks to experiments outcome.

\subsection{Experiments on the application of the framework to simulate MS evolution}
\label{sec:experiments-biomedical}	
	
%
In the following, we first introduce the dataset and the preprocessing steps.
Then, we report and discuss obtained results for each specialization introduced in Section~\ref{sec:specializations}.
It is worth noting that we developed also a web tool for supporting experts in carrying out their analyses online through our framework; such tool will be discussed in Section~\ref{sec:webtool}.

}
\subsubsection{Dataset Description and Preprocessing Steps}


{\color{\red}
Data have been obtained from subjects recruited from the MS clinic of Lyon Neurological Hospital (France).
This prospective study was approved by the local ethics committee (CPP Sud-Est IV) and the French national agency for medicine and health products safety (ANSM).
Written informed consent was obtained from all subjects.
Diagnosis and disease course, constituting the ground truth, were established according to the McDonald's criteria~\cite{lublin2014defining,McDonald01}.
}

Structural connectivity matrices were extracted for each subject.
A total of $578$ samples (distributed into the four aforementioned categories as $63$ CIS, $199$ RR, $190$ SP, $126$ PP, respectively) were considered for the experiments overall, and for each sample the corresponding graph $G$ has been extracted as explained in Section~\ref{sub:mri-graphs}.
{\color{\red} Ground truth on the correct MS stage of all the samples is available, as it has been provided by expert physicians.}
Each graph consists of $84$ vertices with an average of $2036.31 \pm 139.19$ edges for the samples in CIS, $1951.25 \pm 235.43$ in RR, $1634.56 \pm 315.27$ in SP and $1760.96 \pm 293.58$ in PP.

\label{sec:training-classification}

{\color{\red}The ANN introduced in Section \ref{sub:NN-spec} has been trained before starting the experiments on the framework, by cross validation with $3$ folds, using 70\% of the samples in each fold for training the model and the remaining 30\% as test set for validation.}
The quality of the classification was evaluated by means of the average Precision, Recall and F-Measure achieved during the cross validation{\color{\red}, as usual in the literature.}
The proposed ANN was trained using Adam~\cite{DBLP:journals/corr/KingmaB14} with learning rate $0.001$.
Early Stopping was used to prevent overfitting.
Average evaluation of the cross validation is shown in Table~\ref{tab:resultsann}.
It can be observed how the ANN we designed for this work is particularly effective in determining the right stage of the pathology under consideration.
This is a crucial factor in the framework, as studying the impact of the variations in the connectome on the course of the disease requires a very high precision in the classification step. {\color{\red} Observe that, even if augmenting the classification accuracy is beyond the scope of this work, the use of well-performing networks allows to obtain more reliable results.}

Notably, the new classification model adopted in this paper allows to reach an average F-Measure of $88\%$, which represents a significant improvement with respect to the $80\%$ reached in~\cite{DBLP:conf/esann/CalimeriMST18} for the same quality measure.
{\color{\red}It is important to point out that results shown in Table~\ref{tab:resultsann} are slightly lower than previous results obtained in~\cite{DBLP:conf/ruleml/CalimeriCMST18}; nevertheless, we consider these slightly lower values definitely acceptable, as the new ANN used in the present work allows a meaningful and more interpretable representation of the classification process, and this is particularly useful in the new version of the framework.
Indeed, BrainNetCNN leverages the topological locality of structural brain networks, thus performing more meaningful operations on the graph structure with respect to the previous approach~\cite{DBLP:conf/ruleml/CalimeriCMST18} and allowing to compute edge importance.
Furthermore, high-level features learned by BrainNetCNN have been already discussed in the literature in the context of the anatomy and function of the developing pre-term infant brain.

{\color{\red}
After the training phase, in order to keep experiments on the overall framework coherent, before starting the tests on the framework we filtered out some input samples, relying on the ground truth provided by physicians.
In particular, we filtered out input samples misclassified by the trained ANN  at the very first classification; this way, we avoid to propagate initial classification errors in the framework.
As a consequence, a total of $55$ CIS, $189$ RR, $187$ SP and $109$ PP correctly classified samples have been actually fed as input to the framework for the tests.}
}

\begin{table}
\caption{Average Precision, Recall and F-Measure ($\pm$ standard deviation) achieved during cross validation (3 folds). Results are computed per class (CIS, PP, RR, SP) and with respect to all the classes (Tot).}
\label{tab:resultsann}
\begin{tabular}{cccc}
\toprule{}
{}  &     Precision &        Recall &      F-Measure \\
\midrule{}
CIS       &  0.76 ($\pm$0.12) &  0.88 ($\pm$0.13)  &   0.81 ($\pm$0.10)  \\
PP         &  0.91 ($\pm$0.04) &  0.69 ($\pm$0.17)   &   0.78 ($\pm$0.10) \\
RR        &  0.89 ($\pm$0.03) &  0.94 ($\pm$0.06)   &  0.91 ($\pm$0.02) \\
SP        &   0.90 ($\pm$0.06) &  0.93 ($\pm$0.02)  &  0.92 ($\pm$0.03)  \\ \hline
Tot &  0.89 ($\pm$0.01) &  0.88 ($\pm$0.01) &  0.88 ($\pm$0.02)    \\
\bottomrule{}
\end{tabular}
\end{table}


\subsubsection{Experiments for studying structural properties}
In the context of this analysis, we are interested in studying the possible variations of each stage of the MS clinical course by modifying the connectome of a patient, according to the structural properties introduced in Section~\ref{sub:structural-spec}.
For the sake of presentation and space constraint, we discuss in detail the results of a subset of the experiments we carried out, namely {\em Max Clique}, {\em Min Vertex-Cover}, and {\em k-hub}, with $p=50\%$.
Nevertheless, the complete set of results is available at~\additionalmaterial.

Figures~\ref{fig:clique_normal_random}, \ref{fig:vertex-cover_normal_random}, and~\ref{fig:k-hub_normal_random} report the overall results for this specialization. For each starting stage of the pathology, we report the probability values (indicated by a group of four vertical bars, for each iteration of the framework),
computed by the ANN. In particular, from left to right, one can observe the variation of the average probability values for each class. As an example, the first bar in the leftmost group of the first bar chart in Figure~\ref{fig:clique_normal_random} represents the probability associated with the CIS stage for a CIS classified patient. The same bar in subsequent groups shows variations of this probability through iterations 1--4 with the ASP program shown in Figure \ref{alg:clique}; the same bar in the chart below shows variations of the same probability for patients formerly classified as RR.

{\color{\red} In order to evaluate the significance of obtained results, we considered also a \emph{random test} for each test case, designed as follows: at each iteration, given the number $n$ of edges identified by the \emph{Brain Evolution Simulation} module, we generate a parallel modified graph choosing $n$ {\em random} edges to be modified that are not related to the structure under consideration.
}
In other words, we are interested in evaluating whether the variations in classification results depend on the structure of the modified portion of the connectome, or simply on the number of varied edges.
Outcomes of random tests are reported on the right side of Figures~\ref{fig:clique_normal_random}, \ref{fig:vertex-cover_normal_random}, and~\ref{fig:k-hub_normal_random}.

Results show interesting variations, when testing the framework with the \emph{Max Clique} criterion (Figure~\ref{fig:clique_normal_random}).
In particular, it is worth noting that \emph{Max Clique} seems to affect mostly the CIS stage, as CIS probability values significantly decrease.
Interestingly, this behaviour seems not to be simply related to the amount of modified edges: random tests show substantially constant probability values across iterations.
More interestingly, the aforementioned behaviour is not observable for the other stages RR, SP and PP, where the alteration of cliques does not actually induce significant changes.
This absence of variations is not related to the absence of cliques to change, or to their different cardinalities; indeed, the number of edges modified in all stages are comparable being on average $258.37 \pm 34.30$ from iteration $1$ to iteration $2$, and $130.38 \pm 5.34$ from iteration $3$ to iteration 4.
The results for the CIS starting stage also show that probabilities of PP actually increase through iterations, even if not sufficiently enough to allow a guess over a change of state.

As far as \emph{Min Vertex Cover} is concerned (Figure \ref{fig:vertex-cover_normal_random}), significant variations can be observed in all stages.
However, we observe in this case also a significant decrease of the probability of election in random tests, especially in the first iterations.
This is mainly due to the high number of modified edges, being on average $1490.45 \pm 147.12$ from iteration $1$ to iteration $2$, i.e. 41.59\% of the total. 
In the subsequent iterations, very small minimum vertex cover could be identified in the modified graphs due to the low number of remaining edges; as a consequence, very few edges (about 0.003\% of the total) are modified and, thus, very small variations on probabilities are detected.

However, if we concentrate on non random tests, we observe a quite different and interesting behaviour.
In particular, at iteration $2$ the probability of RR is always the highest suggesting that (the absence of) this sub-structure might characterize the RR class, and calls for further studies.


%

Finally, the \emph{$k$-hub} sub-structure (see Figure~\ref{fig:k-hub_normal_random}) can be considered as a counter-example of previous results.
In fact, even if the number of modified edges is significant and comparable with \emph{Max Clique}, i.e., $301.47 \pm 22,82$ from iteration $1$ to iteration $2$, and $202,99 \pm 26,47$ from iteration $3$ to iteration 4, probability values across iterations are almost constant and very similar to the random tests.
This leads us to hypothesize that $k$-hub sub-structures are not characterizing any stage of the disease.

\begin{figure}[ht]
    \centering
    \subfloat[][\emph{Clique (normal)}]{{
    	\includegraphics[width=.5\textwidth]{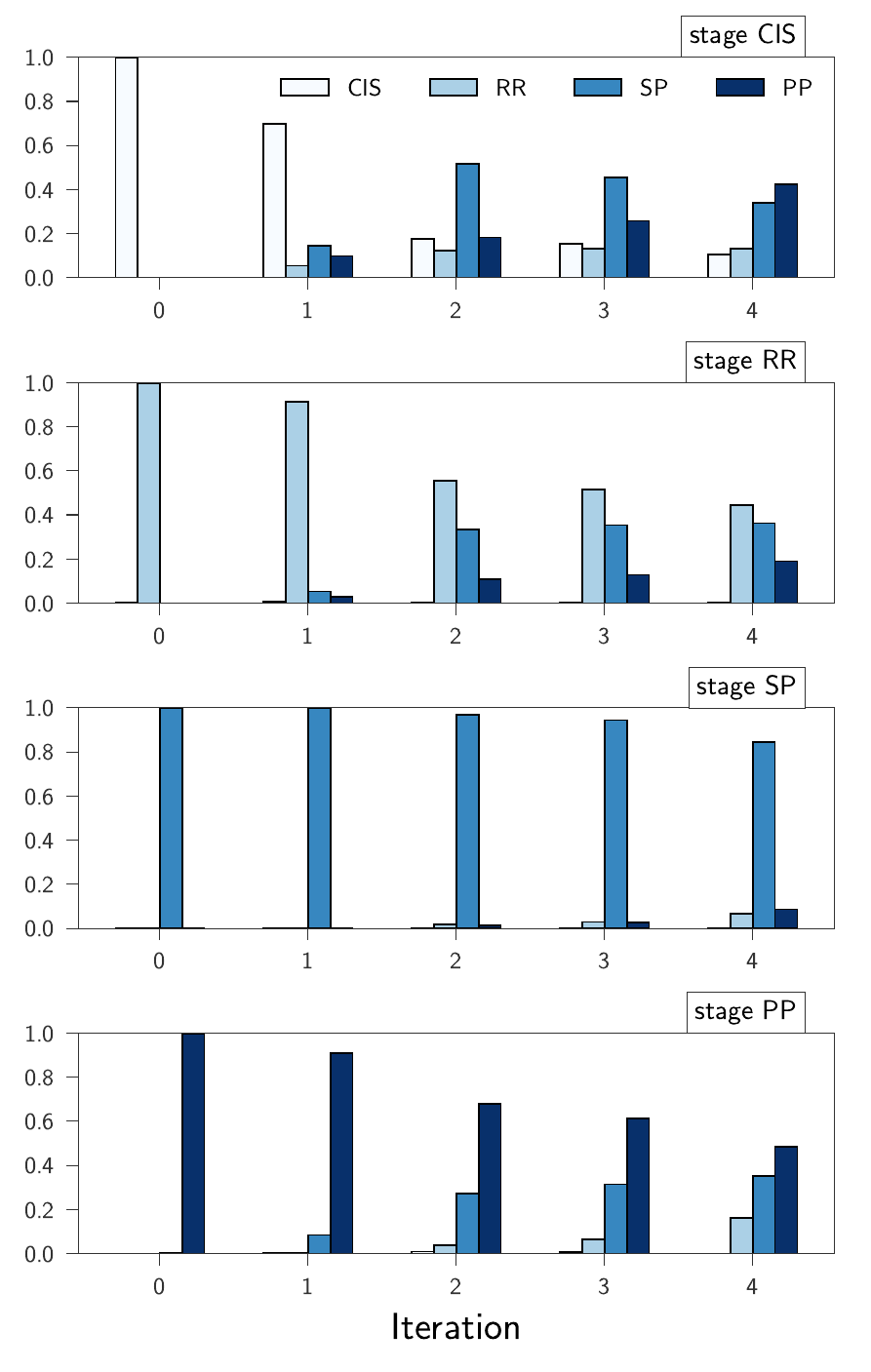}
	}}
    \subfloat[][\emph{Clique (random)}]{{
    	\includegraphics[width=.5\textwidth]{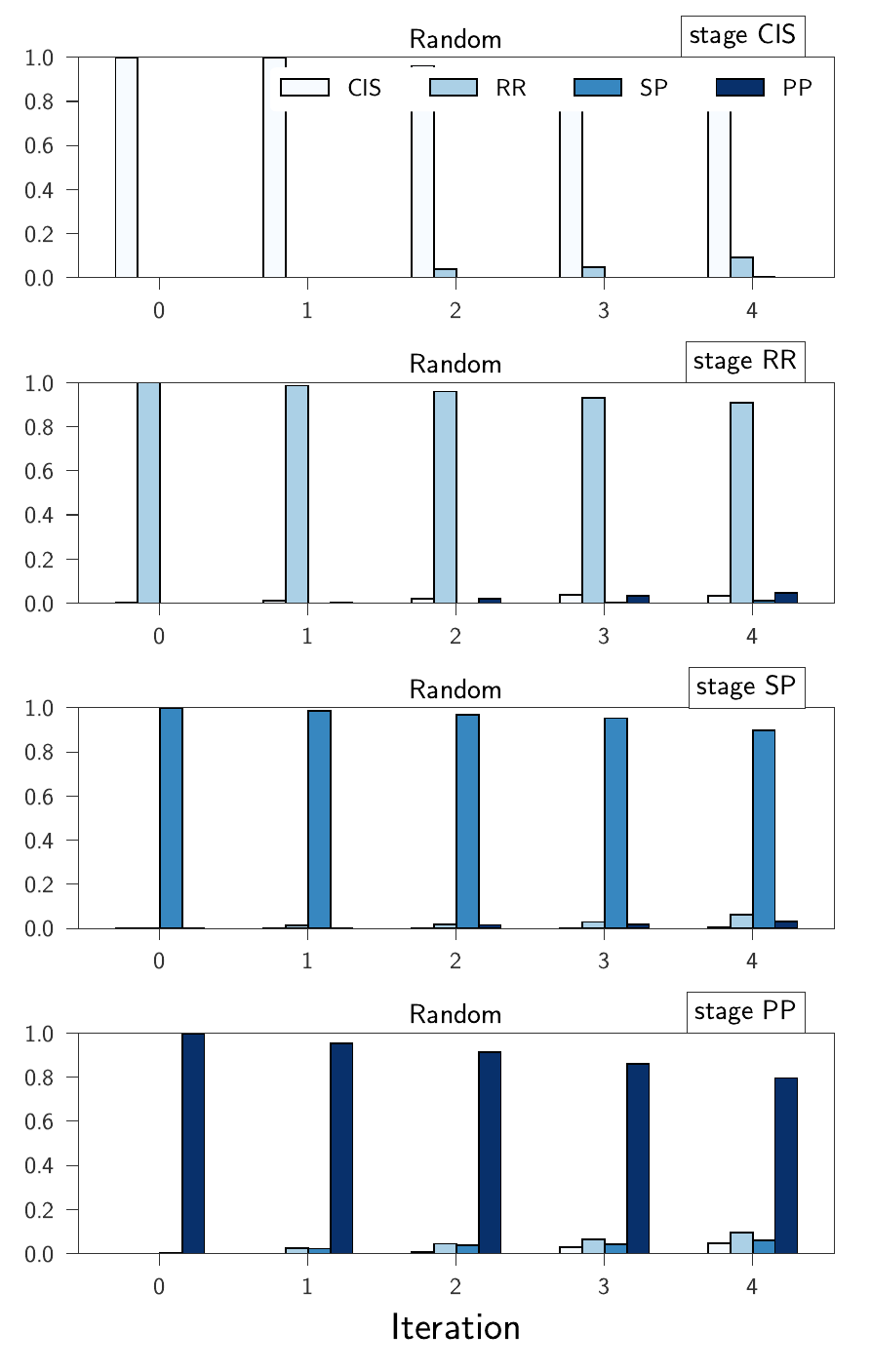}
	}}
 \caption{Results for \emph{Clique} (iterations $i=0..4$).}
    \label{fig:clique_normal_random}
\end{figure}

\begin{figure}[ht]
    \centering
    \subfloat[][\emph{Min Vertex-Cover (normal)}]{{
    	\includegraphics[width=.5\textwidth]{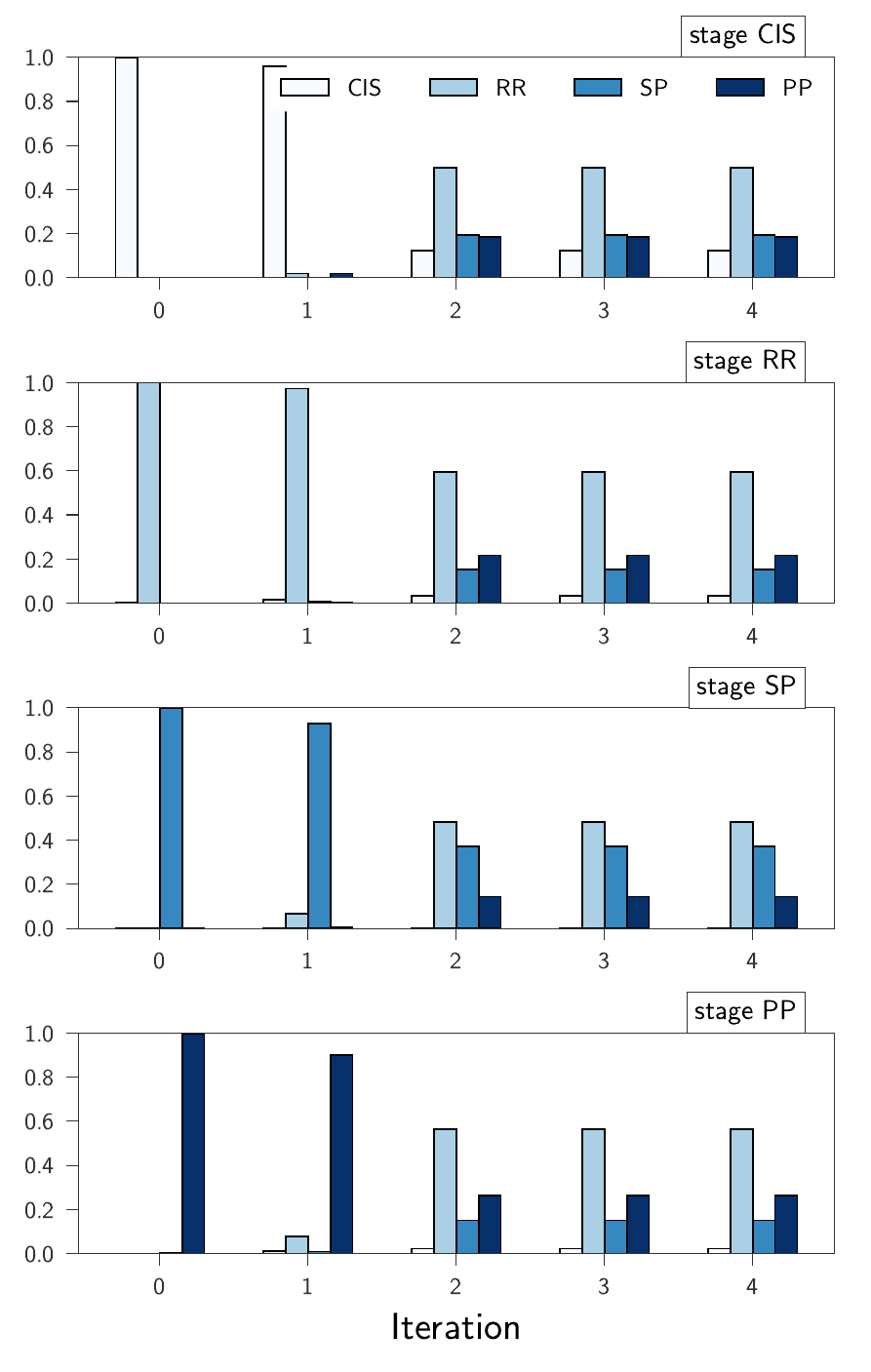}
	}}
    \subfloat[][\emph{Min Vertex-Cover (random)}]{{
    	\includegraphics[width=.5\textwidth]{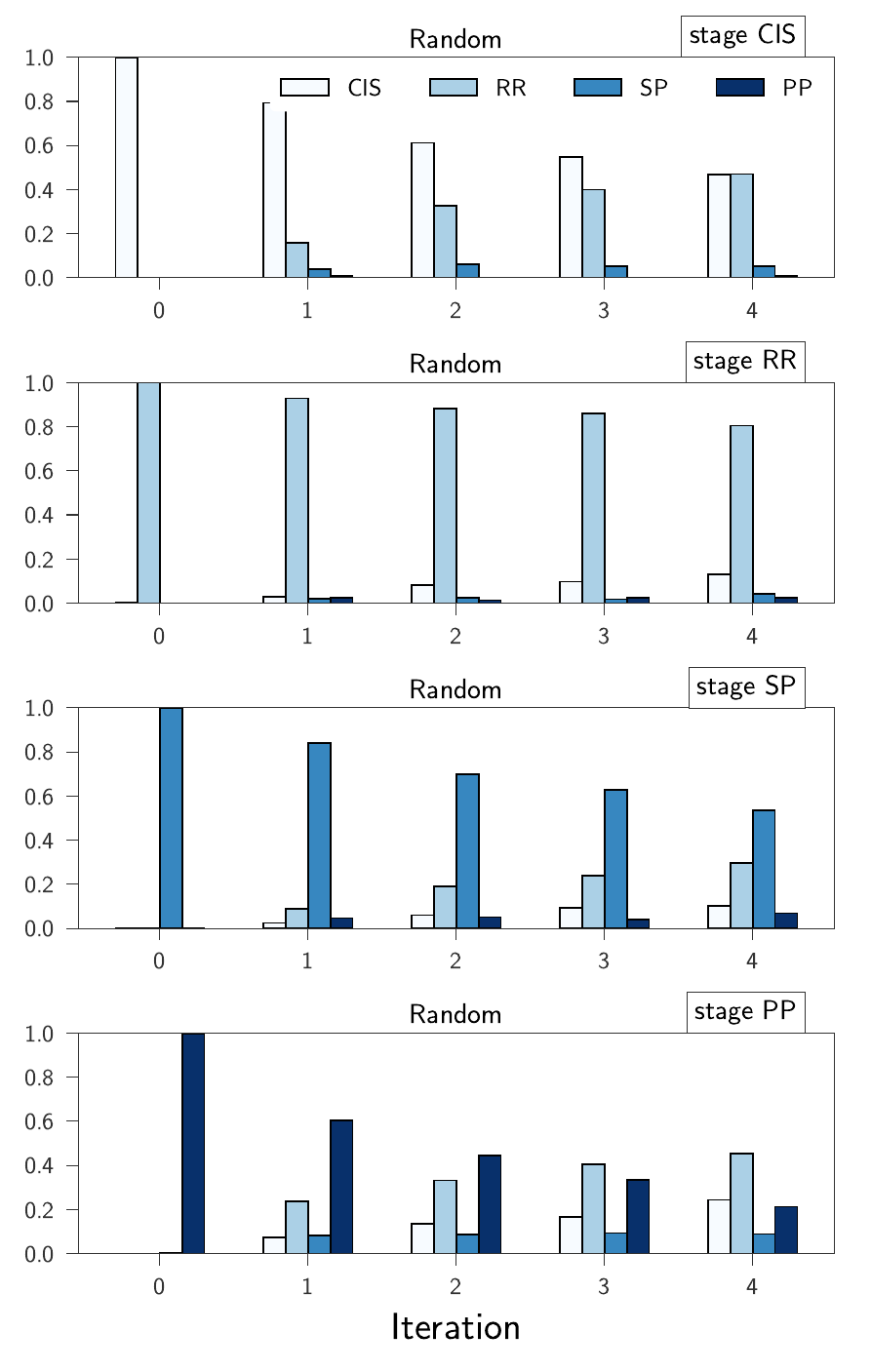}
	}}
 \caption{Results for \emph{Min Vertex-Cover} (iterations $i=0..4$).}
    \label{fig:vertex-cover_normal_random}
\end{figure}

\begin{figure}[ht!]
    \centering
    \subfloat[][\emph{$k$-hub (normal)}]{{
    	\includegraphics[width=.5\textwidth]{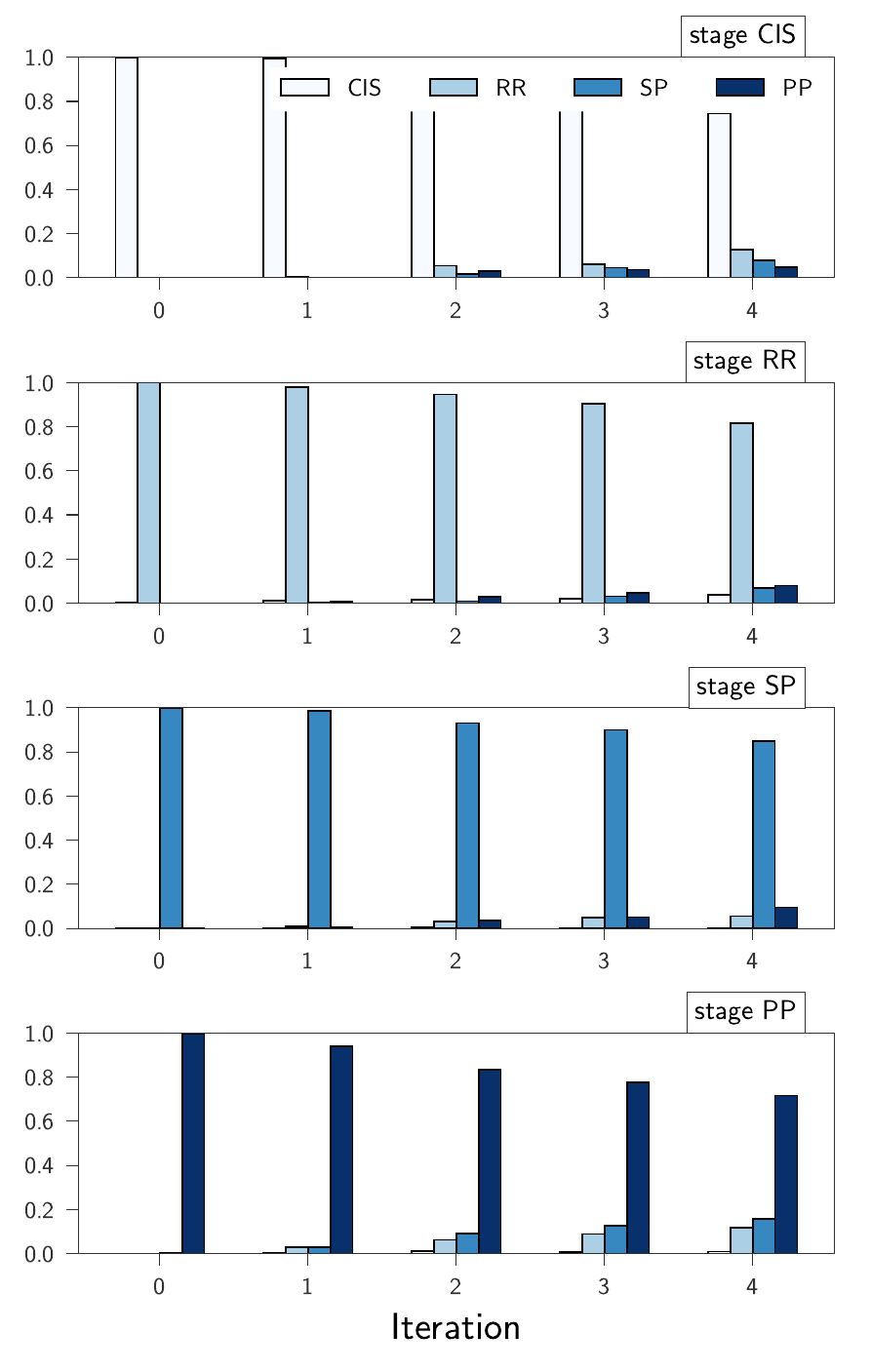}
	}}
    \subfloat[][\emph{$k$-hub (random)}]{{
    	\includegraphics[width=.5\textwidth]{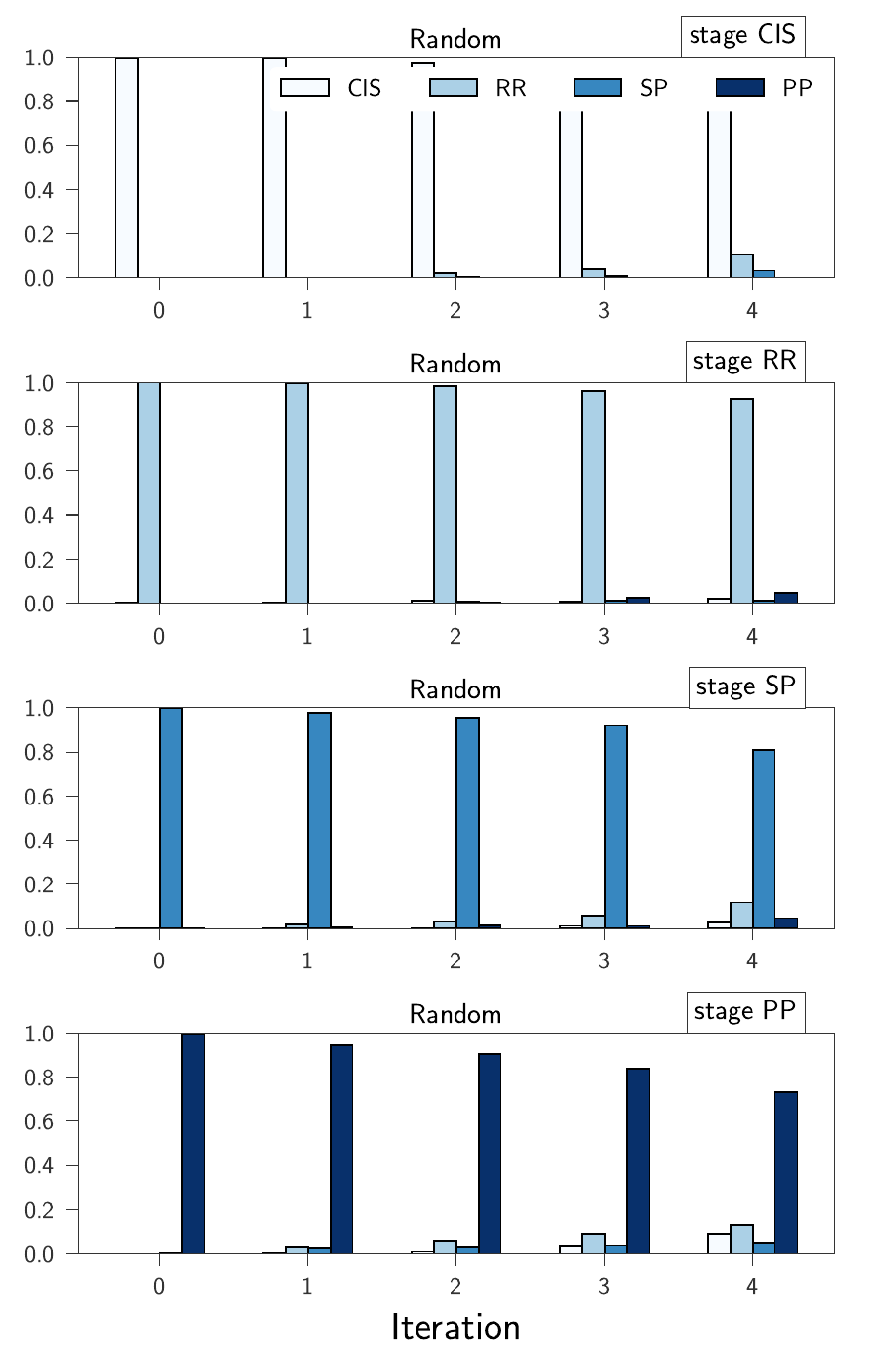}
	}}
 \caption{Results for \emph{$k$-hub} (iterations $i=0..4$).}
    \label{fig:k-hub_normal_random}
\end{figure}

\subsubsection{Experiments for studying graph metrics}
\label{sec:experimentsmetrics}

In the context of this analysis, we are interested in studying the possible variations of each stage of the MS clinical course by modifying the connectome of a patient, according to the metrics introduced in Section~\ref{sub:metrics-spec}. In particular, in this section, we discuss the results obtained for {\em Density} and {\em Assortativity}.
It is worth pointing out that current versions of state-of-the-art ASP systems have not been able to reasonably scale over the graphs involved in the following tests; as a consequence, after proving the viability of the approach over small examples, we simulated the behavior of the ASP-based module via ad-hoc heuristic algorithms.

Figure \ref{fig:density} shows the results obtained for {\em Density}; recall that, at each iteration, we reduce the density of each graph by 10\% by removing edges.
From the analysis of Figure \ref{fig:density}(a) it is possible to observe a generalized reduction of the probability of the initial classification through the iterations. Only for the CIS stage, however, this is more remarkable, even if a clear transition {\color{\red} cannot} be observed from this figure. In order to better analyze this result, we computed (see Figure \ref{fig:density}(b)) the average probabilities obtained by isolating the cases in which a transition from CIS to PP is observed through the iterations (these are 39.53\% of the total) and the cases in which a transition from CIS to RR is observed through the iterations (these are 62.79\% of the total). We point out that some of the graphs presented both transitions through the iterations; they have been considered in both cases. The analysis of Figure \ref{fig:density}(b) shows indeed that there is quite an interesting evidence of these transitions in the last iterations; first of all, these results partially confirm the results obtained from studies presented in~\cite{graphbased}, pointing out that the {\em Density} of a graph characterizes the stage of MS. Moreover, these results provide more insight in the potential \emph{evolution} of the disease. In fact, depending on the area of the brain that is altered, the transition from CIS can evolve in RR stage. This calls for further studies on this aspect.

In order to better analyze these results, we computed the average number of edges removed at each iteration for the two transitions separately (CIS-PP and CIS-RR). Results, reported in Table \ref{tab:edges_removed_cis_density}, show that even a low number of removed edges, if properly selected, may induce a significant change in MS stage.

\begin{figure}[ht!]
	\centering
	\subfloat[][\emph{Results for all stages}]{{
			\includegraphics[width=.5\textwidth]{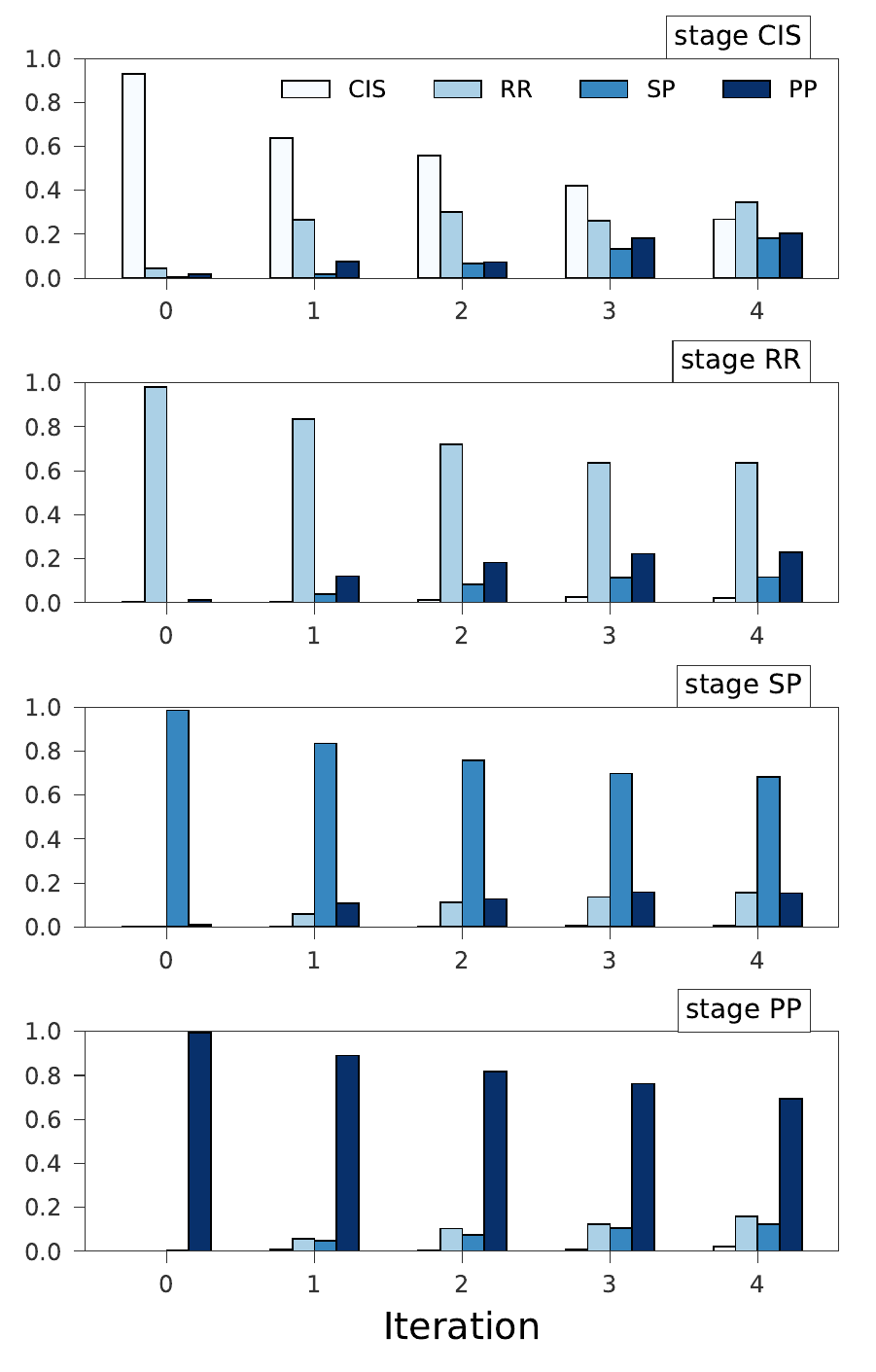}
	}}
	\subfloat[][\emph{Detail for CIS transitions}]{{
			\raisebox{0.5\height}{
			\includegraphics[width=.5\textwidth]{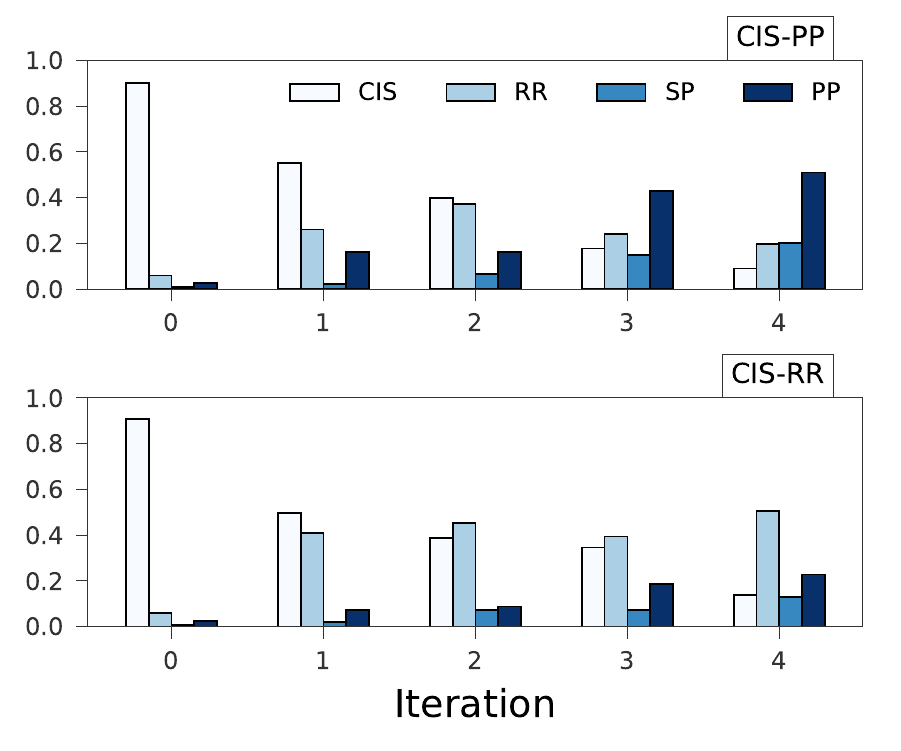}}
	}}
	\caption{Results for
		\emph{Density} (iterations $i=0..4$).}
	\label{fig:density}
\end{figure}

Results obtained for {\em Assortativity} are shown in Figure \ref{fig:assortativity}. Recall that, in this case, we modified the graphs in order to obtain an increase in the property of 10\% at each iteration by removing edges. As far as these results are concerned, we observe an almost complete independence of the computed probabilities on this property through the iterations.
Observe that the final variation of assortativity at the last iteration is about 40\% of its initial value; as a consequence, this result {\color{\red} cannot} be motivated by a low variation in the property itself. As a matter of fact, the number of edges to be removed in order to reach the variation goal on assortativity was indeed extremely small. As an example, only $27.67 \pm 15.94$ edges have been removed on average through the four iterations for CIS patients, and only $24.26 \pm 16.07$ edges for PP patients.
This result can be linked to the ones obtained for \emph{Density} where the specificity of removed edges is probably more important than the overall properties of the corresponding graphs. And this calls for a deeper analysis on the role of specific subsets of edges in the classification process, which is precisely what we analyze in the next section.

\begin{figure}[ht!]
	\centering
	\subfloat[][\emph{Assortativity}]{{
			\includegraphics[width=.5\textwidth]{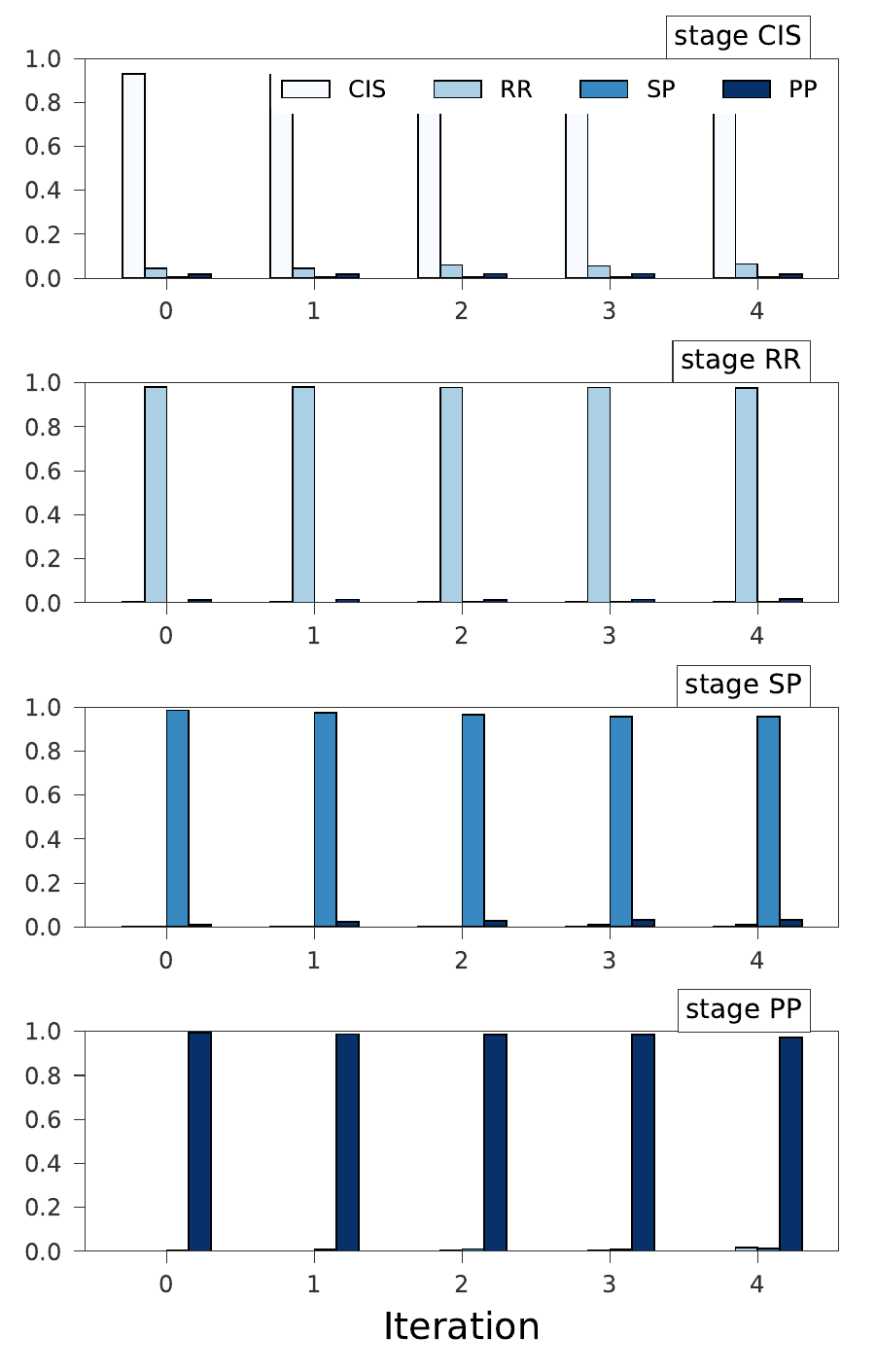}
	}}
	\caption{Results for
		\emph{Assortativity} (iterations $i=0..4$).}
	\label{fig:assortativity}
\end{figure}

\subsubsection{Experiments for studying ANN insights}
\label{sec:experiment-insights}

\begin{table}[t!]
	\caption{Average number of removed edges ($\pm$ standard deviation) for reducing \emph{Density} of 10\% at each iteration.}
	\label{tab:edges_removed_cis_density}
	\begin{tabular}{ccc}
		\toprule
		Iteration  &     CIS-PP &        CIS-RR   \\
		\midrule
		1       &  190.65 ($\pm$15.82) &  199.48 ($\pm$11.15)  \\
		2         &  171.47 ($\pm$14.19) &  179.52 ($\pm$9.96)  \\
		3        &  154.47 ($\pm$12.87) &  161.67 ($\pm$9.00)   \\
		4        &   138.94 ($\pm$11.48) &  145.41 ($\pm$8.14)  \\ \hline
		Tot &  655.53 ($\pm$54.35) &  686.07 ($\pm$38.23) \\
		\bottomrule
	\end{tabular}
\end{table}

In Section \ref{sub:insights-spec} we introduced the concept of importance of an edge for the classification purposes, computed by exploiting the peculiar properties of the adopted ANN. In particular, we introduced a specialization of our framework dealing with edge importance.

Before analyzing our tests coupling graph structures or graph metrics with edge importance, we first consider the impact of the importance degree through the following two simple tests. Assume edges are ordered by importance, in descending order: {\em (i)} remove, one by one, the edges starting from the most important ones; {\em (ii)} remove, one by one, the edges starting from the less important ones. Each time an edge is removed, the classification task is carried out again and results plotted for each stage. Results are shown in Figure \ref{fig:just-importance}. Consider the results in Figure \ref{fig:just-importance}(b) first; from the analysis of this figure, it is manifest that removing unimportant edges is completely irrelevant for the classification result at any stage. And this is true up to about 1500 removed edges. On the contrary, removing even very few important edges may strongly affect classification results (see Figure \ref{fig:just-importance}(a)).

These considerations call for a deeper analysis of the two settings tested above. In particular, it would be of great interest for an expert both knowing that there are edges he/she may completely disregard in the analysis, and that there are edges that need more attention for stage variation analyses.

In particular, given the specializations for the Brain Evolution Simulation module introduced in Section \ref{sub:insights-spec}, we set a threshold such that the top 40\% of edges having the highest importance are considered as important, whereas the remaining edges are considered non important.

Figure~\ref{fig:analysis-with-importance}(a) shows results for \emph{Max Clique} when unimportant edges only are allowed in the cliques.
The number of altered edges is comparable to the one obtained for the previous tests on \emph{Max Clique}.
Figure~\ref{fig:analysis-with-importance}(a) clearly confirms that altering graph sub-structures using unimportant edges only, provides no apparent modifications in the classification.
As a consequence, it confirms the fact that experts may completely disregard at least 60\% of edges in their analyses.

On the contrary, Figure \ref{fig:analysis-with-importance}(b) reports results on \emph{Density} when only important edges are removed. In this case, results show that when a huge amount of important edges {\color{\red} is} removed, the ANN becomes almost unable to perform a reliable classification. As a consequence, this reinforces the intuition that there are very few important, and let's say \emph{critical}, edges guiding transitions between MS stages. In Section \ref{sec:webtool} we show how we took these preliminary results into account in order to provide experts with a powerful analysis tool.

\begin{figure}[ht!]
	\centering
	\subfloat[][\emph{Starting from the important edges}]{{
			\includegraphics[width=.5\textwidth]{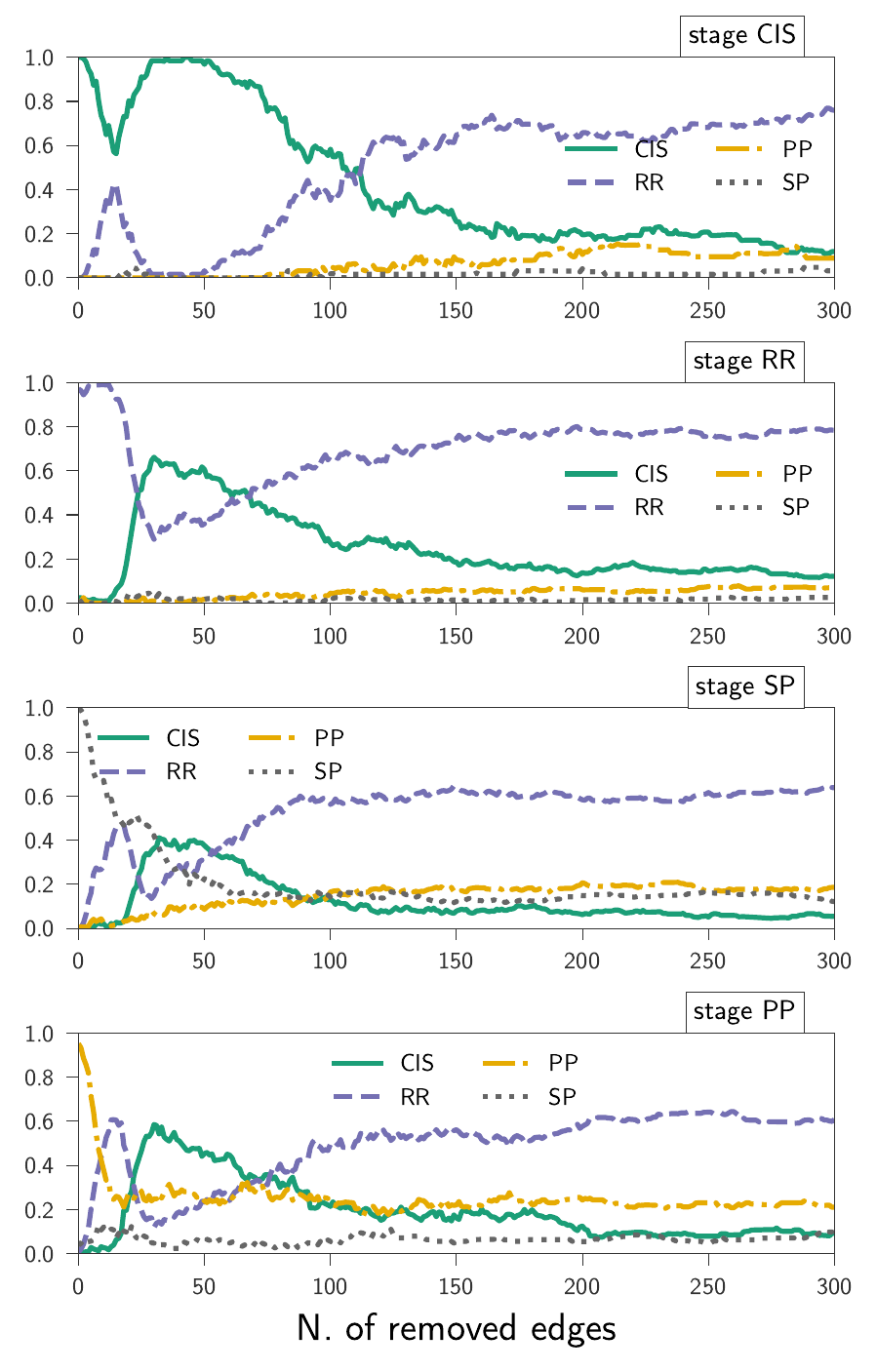}
	}}
	\subfloat[][\emph{Starting from the less important edges}]{{
			\includegraphics[width=.5\textwidth]{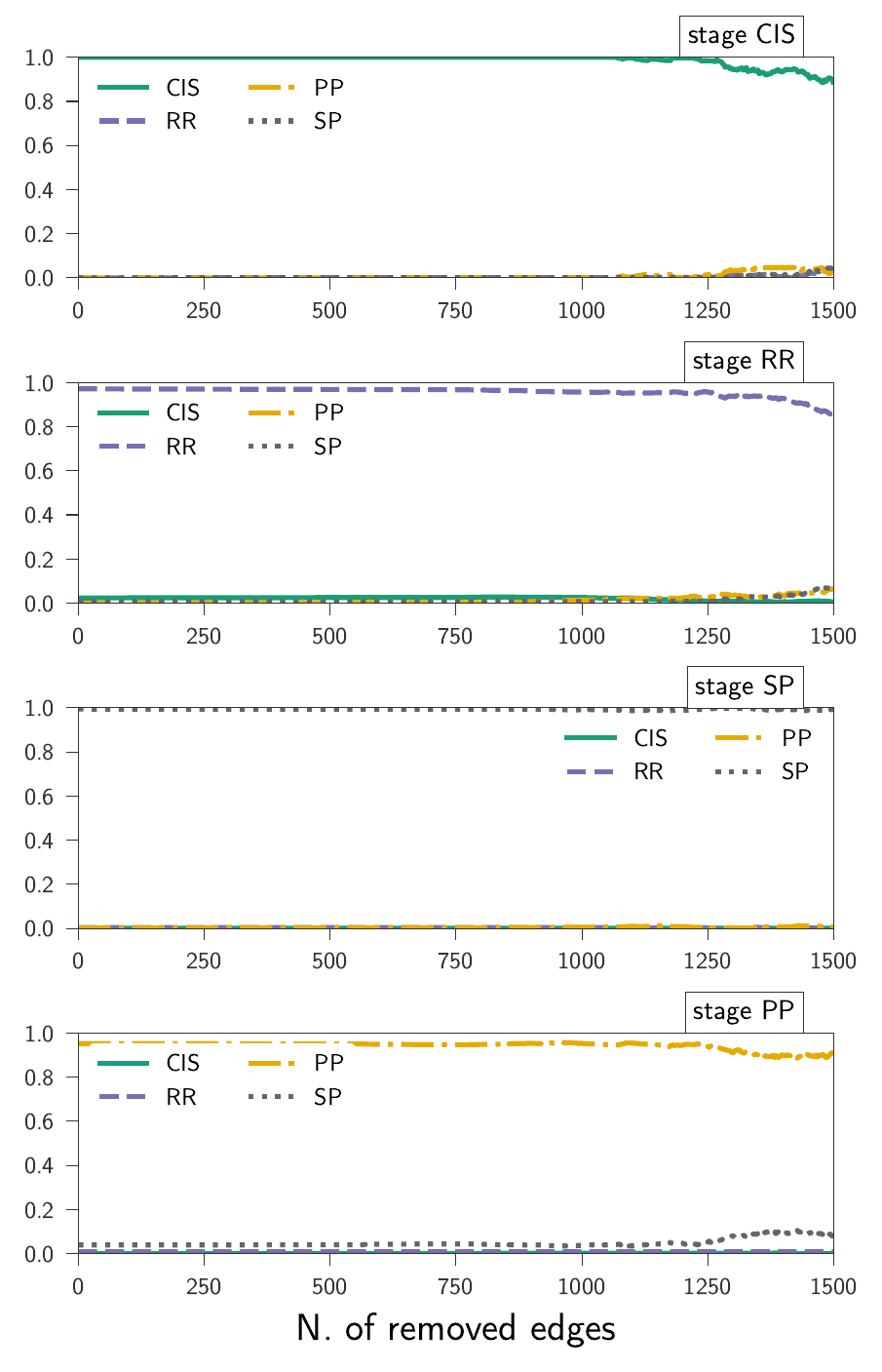}
	}}
	\caption{Variation of classification results removing important/unimportant edges.}
	\label{fig:just-importance}
\end{figure}

\begin{figure}[ht!]
	\centering
	\subfloat[][\emph{Clique - Altering unimportant edges}]{{
			\includegraphics[width=.5\textwidth]{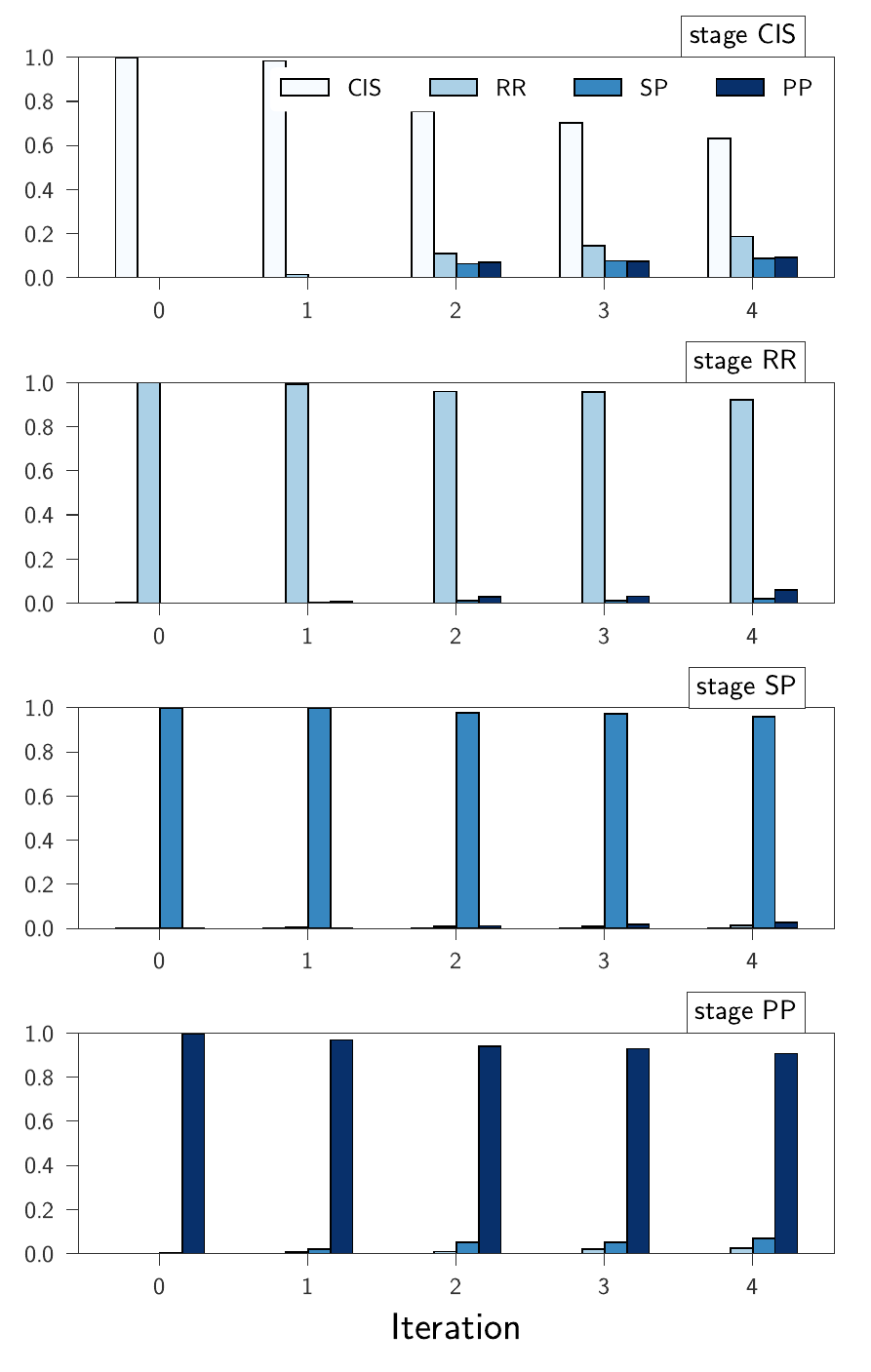}
	}}
	\subfloat[][\emph{Density - Removing important edges}]{{
			\includegraphics[width=.5\textwidth]{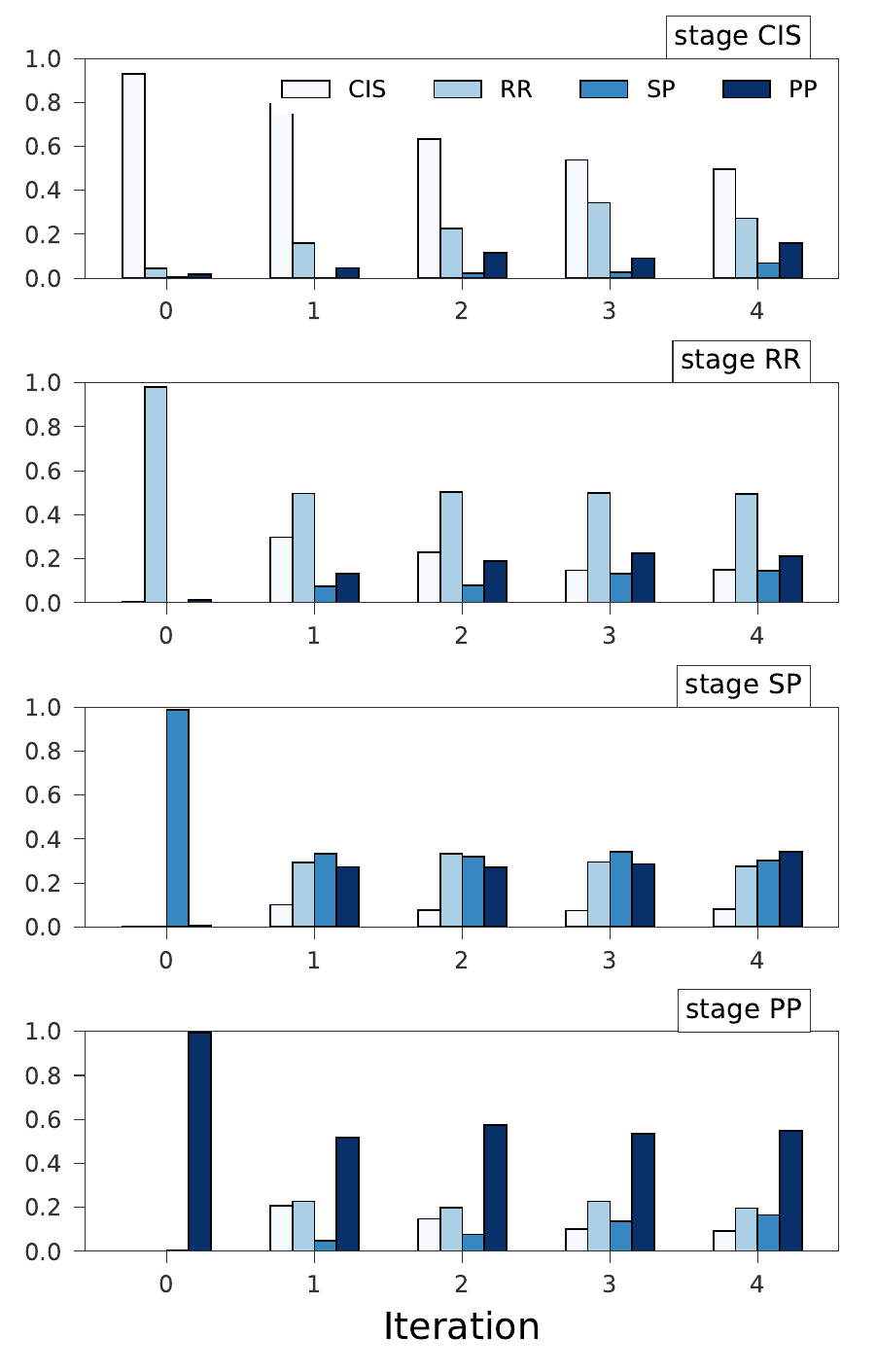}
	}}
	\caption{Analyzing structural properties and graph metrics considering important/unimportant edges.\label{fig:analysis-with-importance}}
\end{figure}

{\color{\red}

\subsection{Experiments on performances}
\label{sec:performances}	

In this section we present the results of a series of tests aimed at providing support for
a performance analysis of the system.
We first analyze execution times of  the entire framework, then we single out the role of the main modules in the overall execution times.
Eventually, we focus on the ASP-based \emph{Brain Evolution Simulation} module.

All tests have been carried out on a Linux machine $4.15.0-20$-generic \#21-Ubuntu, with an Intel(R) Core(TM) i$7$-$4770$ CPU @$3.40$Ghz and $15.6$ GB of RAM. As for the grounder and the solver, we coupled I-DLV (version 1.1.0) and WASP (version 2.0). The solver parameters have been set to \texttt{--silent} and \texttt{--printonlyoptimum}; this means, that the computation is stopped at the first optimum found for optimization problems.

When not differently specified, we used the dataset introduced in Section~\ref{sec:training-classification}.
It is worth pointing out that the aim of these tests is not an assessment of ASP solvers and their performance (as extensively done in related literature~\cite{DBLP:journals/aim/CalimeriIKR12,DBLP:conf/aaai/GebserMR16,DBLP:journals/jair/GebserMR17,DBLP:journals/corr/abs-1904-09134}), but rather to assess the applicability of the proposed framework to the herein considered context and related ones.

\begin{figure}[t] \includegraphics[width=1\textwidth]{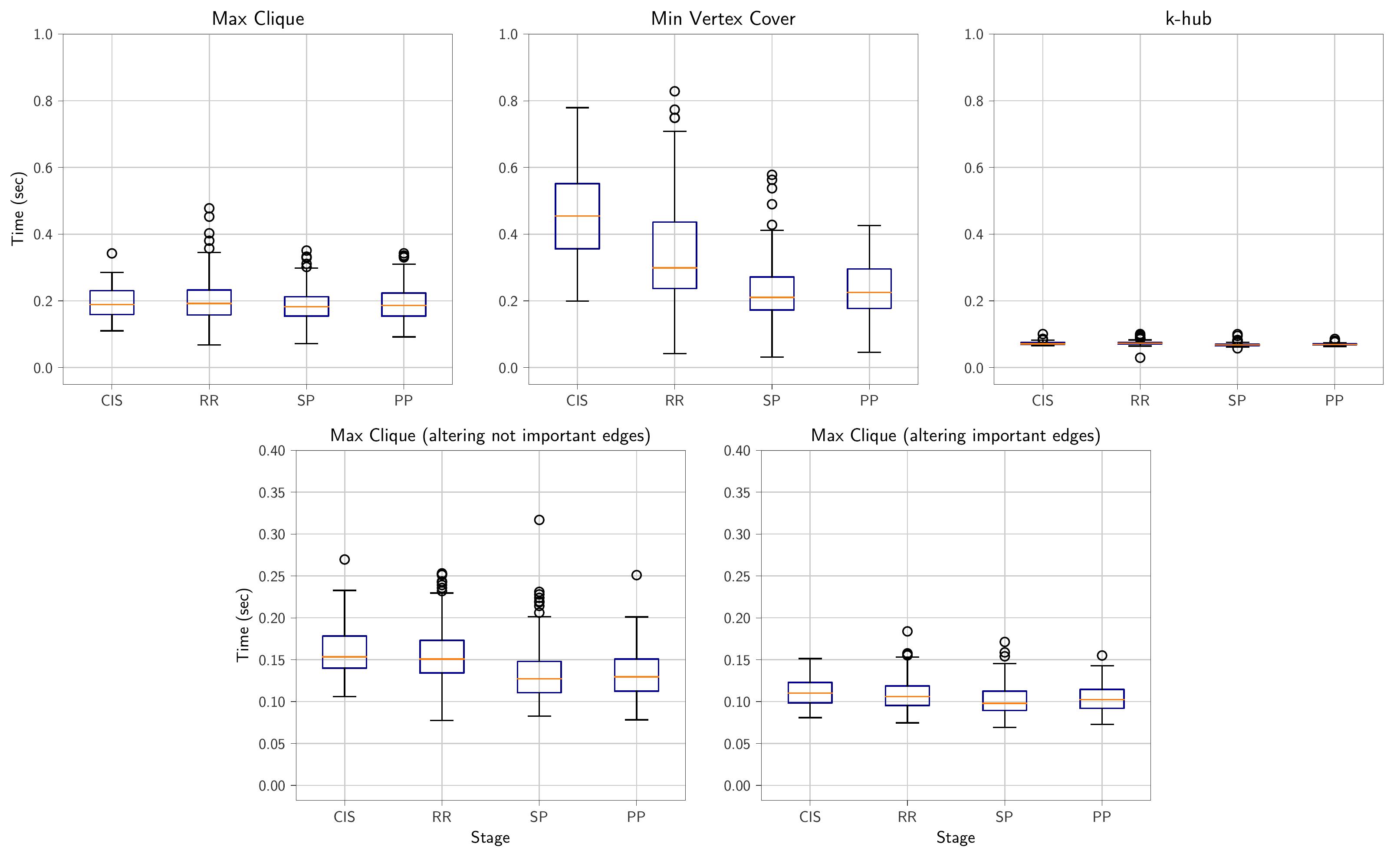}
	\caption{Execution times for one iteration of the framework, considering the three structural properties {\em Max Clique}, {\em Min Vertex-Cover}, and {\em k-hub}.
Bottom graphs show execution times for one iteration of the framework on {\em Max Clique} considering either important or unimportant edges.\label{fig:times-one-iteration}}
\end{figure}

\begin{figure}[h!t] \includegraphics[width=.75\textwidth]{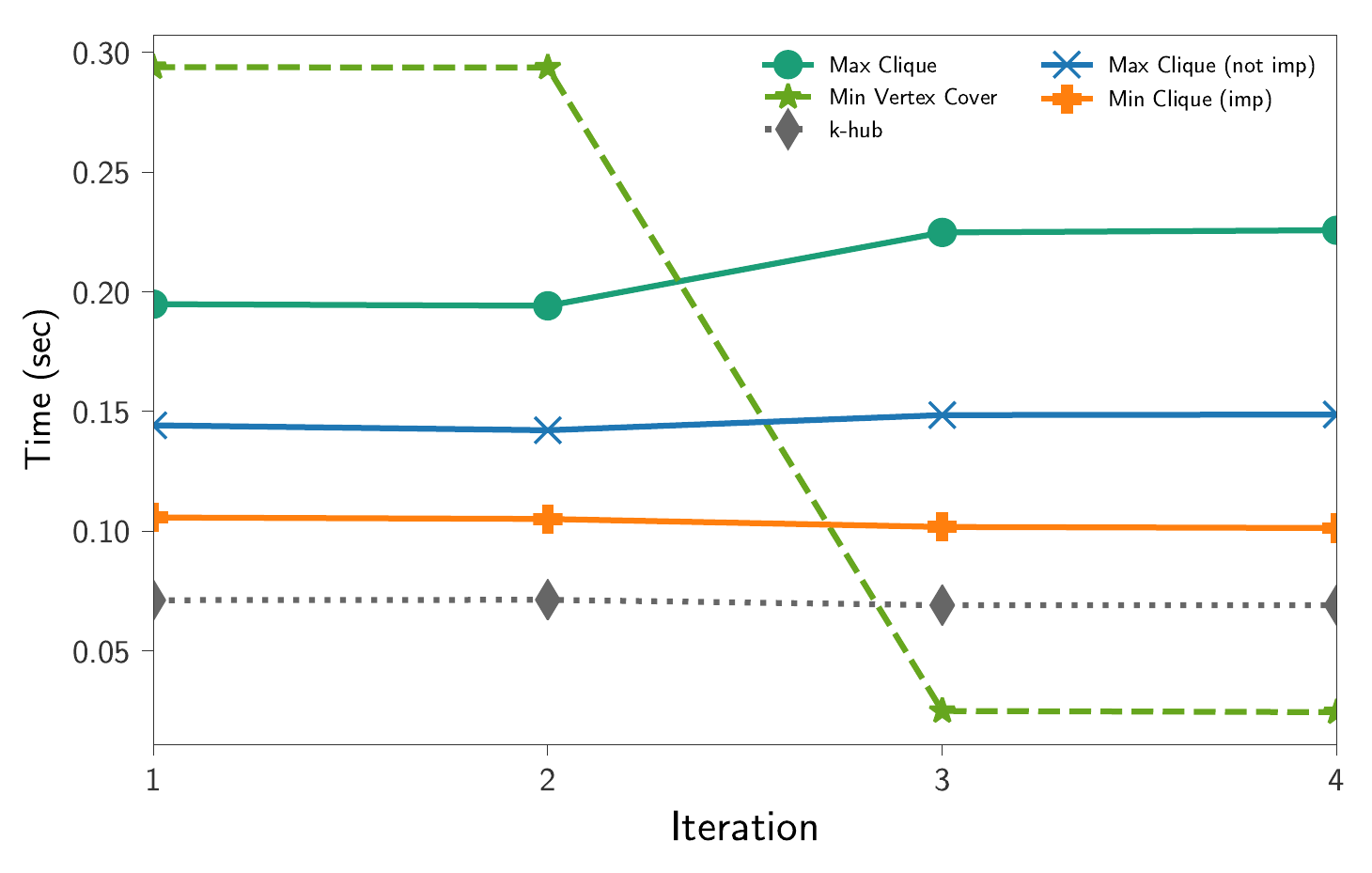}
	\caption{Execution times for four iteration of the framework.\label{fig:times-four-iterations}}
\end{figure}

In a first series of experiments, we measured the execution times of one iteration of the framework, i.e., involving one step in the brain evolution simulation.
We separately considered the three structural properties {\em Max Clique}, {\em Min Vertex-Cover}, and {\em k-hub} discussed above and, as for {\em Max Clique}, we tested also the ASP programs altering both unimportant and important edges\footnote{{\color{\red}Recall that in Section~\ref{sec:experiments-biomedical} we set a threshold such that the top $40$\% of edges having the highest relevance are considered as important, whereas the remaining edges are considered as unimportant. It is worth mentioning again that information about importance of edges is directly provided by the Classification module.}}.
Results are shown in Figure~\ref{fig:times-one-iteration}.
In order to verify whether the starting stage influences performance or not, we highlighted running times for each property and for each stage; times are averaged over all the samples grouped by MS stage and standard deviation is also shown in the figure.
From the analysis of Figure~\ref{fig:times-one-iteration}, we can observe that
\begin {enumerate*} [(a) ]%
\item
    the execution time of one iteration is particularly small for all properties, always significantly lower than one second;
\item
    there is no actual correlation between starting stages and performance;
\item
    interestingly, considering important/unimportant edges reduces average execution times, as the dimension of the graphs the ASP program works on is reduced, in terms of edges.
\end{enumerate*}

We then evaluated the potential impact on running times due to subsequent iterations.
Figure~\ref{fig:times-four-iterations} reports execution times averaged over all the stages for four subsequent iterations; it is easy to see that fluctuations of running time among iterations are negligible except for {\em Min Vertex-Cover} where, as discussed in Section \ref{sec:experiments-biomedical}, at the third iteration a very low number of edges remains in the modified graphs.

In a further series of experiments, we considered the impact of each module of the framework in the running time of one iteration.
We take into account the three main modules, namely the \emph{Classifier}, the \emph{Classifier Validity Checker} and the \emph{Brain Evolution Simulation} modules.
Results shown in Figure~\ref{fig:times-for-module} clearly point out that the main load of computation is on the \emph{Brain Evolution Simulation} module.
Obviously, both \emph{Classifier} and \emph{Classification Validity Checker} execution times are independent from the graph property under examination; interestingly, they are both significantly faster than the simulation task.
Higher execution times for {\em Max Clique} and {\em Min Vertex-Cover} with respect to {\em k-hub} depend on the deterministic nature of the encoding for {\em k-hub}.
The same considerations carried out in the previous tests when including important/unimportant edges for {\em Max Clique} are still valid in this test.

\begin{figure}[ht!]
	\includegraphics[width=1\textwidth]{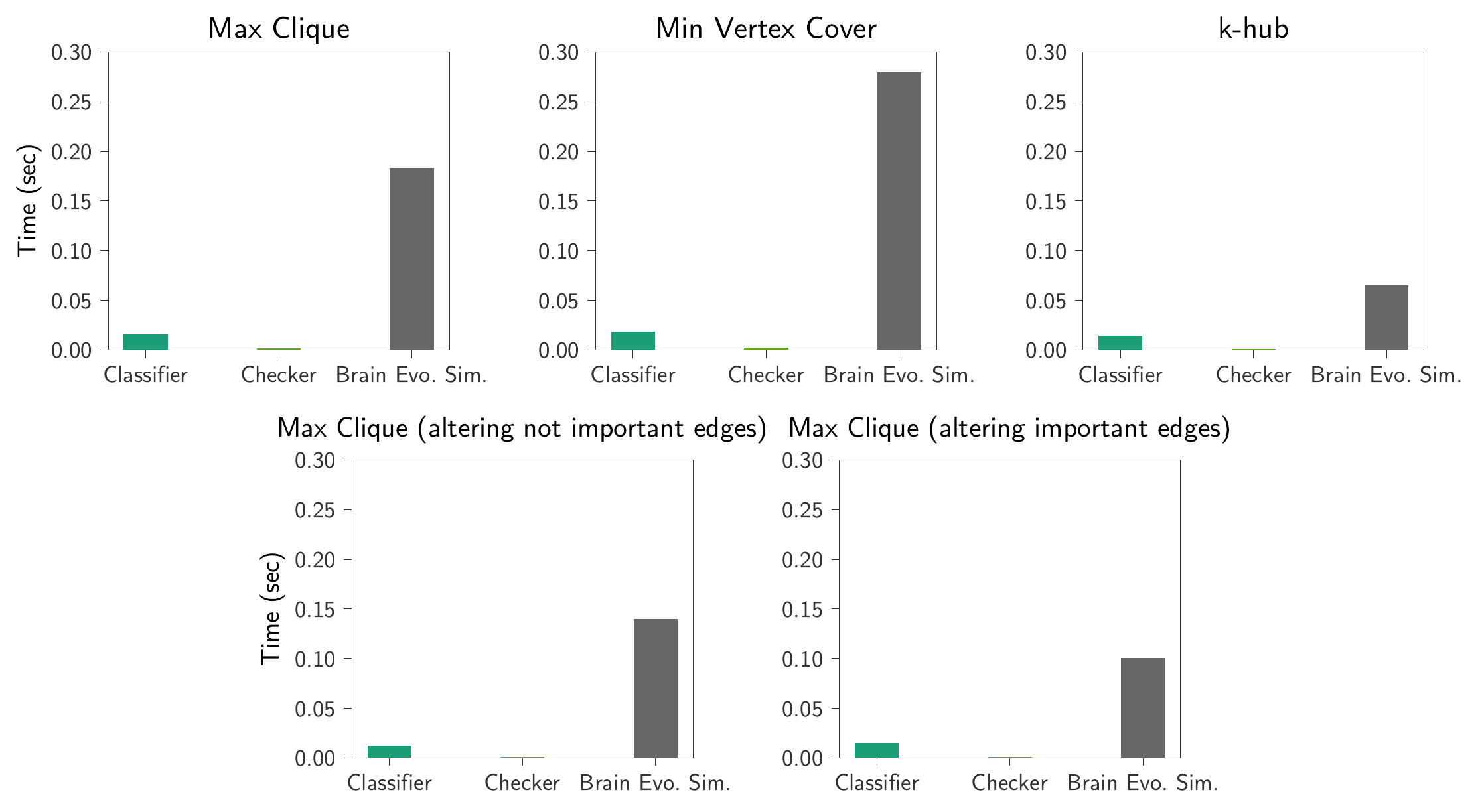}
	\caption{Impact of each module of the framework in the running time of one iteration.}
	\label{fig:times-for-module}
\end{figure}

\smallskip

The scalability of the ASP part of the system has been tested over graphs of increasing size.
First of all, we measured running times of the \emph{Brain Evolution Simulation} module over a set of simulated graphs possibly representing connectome; in particular, we fixed the number of nodes ($84$ in our tests, coherently with the technique described in Section~\ref{sub:mri-graphs}) and we randomly generated graphs with increasing number of edges up to a complete graph. It is worth recalling that, as pointed out in Section~\ref{sec:training-classification}, the average number of edges in graphs corresponding to real connectome is around $2000$.
Results are reported in Figure~\ref{fig:times-for-increasing-edges}; each data point is the average running time of $10$ different executions over random graphs having the same number of nodes and edges.
Via this figure, it is possible to observe that all the tested ASP programs for studying structural properties are solvable with execution times always below one second on any potential graph representing a connectome.
There are obviously small variations between different samples and properties; nonetheless, the figure shows that any connectome can be easily managed by our approach in order to study structural properties.


\smallskip

As a further scalability test, we generated graphs with increasing number of nodes; as for the number of edges in these graphs, we measured the average number of edges in the graphs representing real connectome and we kept the same proportion of edges for each generated graph.
Results are shown in Figure~\ref{fig:times-for-increasing-nodes} for graphs up to 700 nodes; again, each data point is the average running time of $10$ different executions over random graphs having the same number of nodes and edges.
In this case, it is possible to observe that, while the number of nodes is around $100$, running times for all the problems are reasonable; when the number of nodes grows further, the combinatorial explosion of programs including non-deterministic choice rules is reflected in rapidly increasing running times (indeed, as an example, {\em Max Clique} is reported as an hard problem in the ASP Competition series~\cite{DBLP:conf/aaai/GebserMR16,DBLP:journals/jair/GebserMR17,DBLP:journals/corr/abs-1904-09134}); {\em Min Vertex-Cover} is affected first by this issue.
In fact, we observed that, on the machine used, it may require more than one hour of computation for determining the {\em Min Vertex-Cover} on a graph with around $150$ nodes or finding the {\em Max Clique} on a graph with around $500$ nodes.
It is interesting to observe that, again, considering only important/unimportant edges allows to move forward the limit of computation.
Indeed, the lower number of considered edges simplifies the graph and the execution time is faster; as it was expected, considering important edges (40\% of the total) allows a further improvement with respect to the not  important edges (60\% of the total).
Notably, {\em k-hub} scales very well on tested graphs.

\smallskip

These results pose some questions on the applicability of ASP solutions, and generally of exact solutions for optimization problems, in contexts different from the one studied in this paper, where the graphs to be handled become very large and non-deterministic reasoning over the graph is needed. In these contexts, heuristic algorithms, not spanning the entire search space, might be more efficient. However, as previously pointed out, compactness, versatility, and declarative nature of ASP allow for a fast prototyping, and make it an excellent tool for testing numerous alternative graph properties; in those contexts where input is represented by large graphs, one may think of applying ASP based solutions on small sample graphs, in order to identify the most promising properties and, then, implementing them with other ad-hoc, more efficient, solutions in order to study the problem on real graphs. The adoption of ANN insights on important edges shown in this paper may be of significant help in this task; indeed, leveraging the most important edges only allows to work on smaller but still significant graphs.

Furthermore, we do believe that applications like the one herein described can significantly motivate the scientific community, especially the one working on ASP, at improving performance of systems for their use in real-world applications.

\begin{figure}[t]
	\includegraphics[width=.85\textwidth]{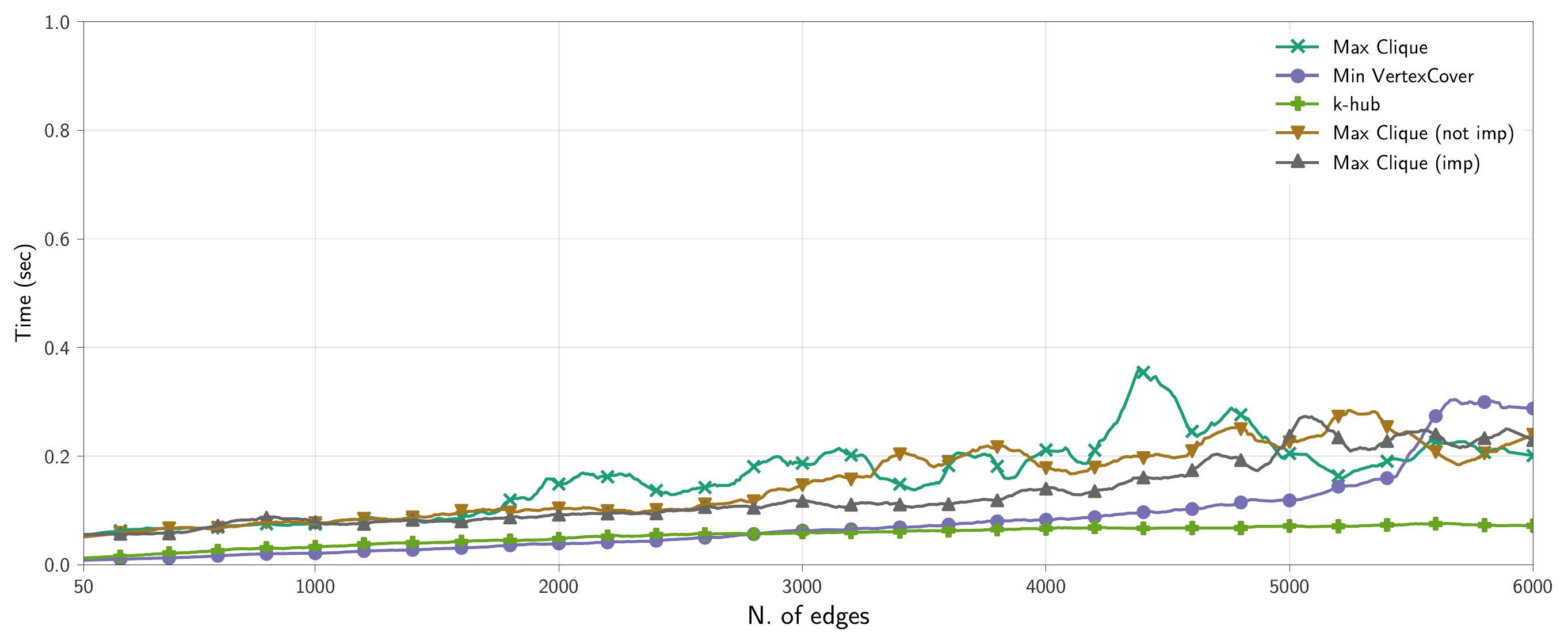}
	\caption{Running time of the Brain Evolution Simulation module on simulated graphs with increasing number of edges.}
	\label{fig:times-for-increasing-edges}
\end{figure}

\begin{figure}[t]
	\includegraphics[width=.85\textwidth]{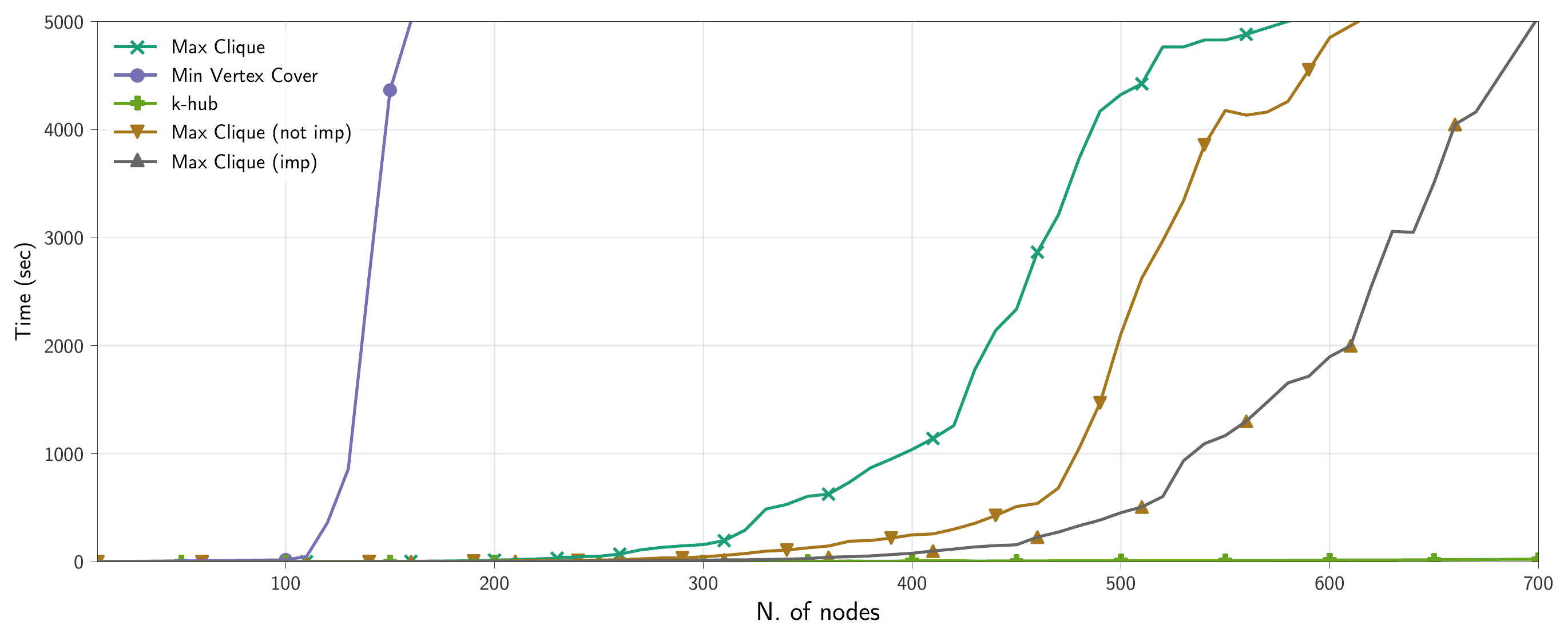}	
	\caption{Running time of the Brain Evolution Simulation module on simulated graphs with increasing number of nodes.}
	\label{fig:times-for-increasing-nodes}
\end{figure}

\subsection{Discussion}\label{sec:discussion}	
The tests presented in this section allow us to draw some interesting considerations.
First of all, all tests presented in Section~\ref{sec:experiments-biomedical} actually proved the appropriateness of the approach for studying the evolution of MS; also, simplicity and high versatility in defining, setting up and carrying out a wide variety of tests showed how crucial is the role played by ASP.
We also provided some experts with the system for a first view, and they have been quite impressed by the possibility of simulating brain evolution so easily.

On the ASP side, we can also say that what discussed in Section~\ref{sec:performances} confirms that expressiveness and compactness of the language make it perfectly suitable to address a wide variety of problems on graphs.
Moreover, language extensions significantly expanded the range of applicability of ASP; as an example, weak constraints allowed us to easily express optimization problems and, analogously, the recently introduced possibility of placing external function calls into a logic program, with a predicate as function parameter instead of a single variable (see, e.g., the encoding in Figure~\ref{alg:decrease}), allowed us to keep the encoding simple and elegant even with the inclusion of complex graph metrics computation.
Furthermore, when the ASP program does not include choice rules, actual ASP implementations can deal with very large graphs, and scale definitely well.
Unfortunately, on the downside, the major weakness of current ASP systems becomes apparent when a combinatorial explosion of the problem occurs.
In particular, while we have shown that the system is fully capable of addressing structural properties on connectome, we experienced that addressing large graphs is possible only to some extent.

Moreover, as pointed out in Section~\ref{sec:experimentsmetrics}, when dealing with graph metrics, since the non-deterministic choice is carried out on edges instead of nodes, and since the metrics need to be computed on the entire guessed graph, current versions of state-of-the-art ASP systems do not reasonably scale over the connectome graphs.
In particular, we observed that systems incur in out-of-memory or exceed time limit (more than one hour) much earlier with respect to the tests focusing on structural graph properties.
Intuitively, the problem is that the systems need to generate all possible guesses on potential graphs, before computing the corresponding metrics.
In order to exclude external function calls as the potential bottleneck in this case, we also checked a version of the program for controlling graph density variations using aggregates only; while avoiding external function calls allowed us to reduce memory issues, we encountered similar scalability issues on connectome.

The experience above calls for the need of some extra features of ASP systems, e.g., extending their solving capabilities with custom heuristics and propagators; some work in this direction is currently ongoing (see~\cite{DoRi18} and references therein).
However, the applicability of these approaches in our context is not straightforward.
As a matter of fact, even taking the possibility of specifying suitable propagators into account, it is not always possible to easily define model generation guiding rules; let us think, for instance, to assortativity, where it is not clear how to guide the edge selection in order to imply a decrease in the property.
Moreover, the problem is even more complex if we consider that the combinatorial explosion of this problem is coupled with an optimization task.

However, it is worth stressing again the potential role of important/unimportant edges in encompassing, to some extent, scalability issues in our general framework.
In fact, we have first shown in Sections~\ref{sec:experiment-insights} (see specifically Figure~\ref{fig:just-importance}) that removing even a high percentage of unimportant edges does not affect the classification quality.
We have then shown in Section~\ref{sec:performances} (see specifically Figure~\ref{fig:times-for-increasing-nodes}) that limiting the computation on (a small number of) important edges only significantly reduces performance issues, thus extending the dimensions of graphs that can be managed.

To the best of our knowledge, this is the first work showing how to exploit the importance of an edge in graph-based classification tasks in order to boost reasoning capabilities over graphs.
In our opinion, this result deserves further investigations in future works, and can stimulate the research community in looking for new ASP program evaluation optimizations.

}} 
\section{Integrated Web Tool}
\label{sec:webtool}

In this section we present an integrated web tool that has been developed in order to implement the framework introduced in this paper and make it actually usable. 
The tool is available online at~\url{https://brainmsa.mat.unical.it}.
The integrated environment provides a user-friendly interface that shows analysis results in real time.
The main objective of the tool is to help physicians, typically neurologists that are not likely to be ANN/ASP experts, to study the evolution of MS through the application of the proposed framework, but also by manual inspection of brain modifications.


\begin{figure}[t]
	\centering
	\includegraphics[width=.8\textwidth]{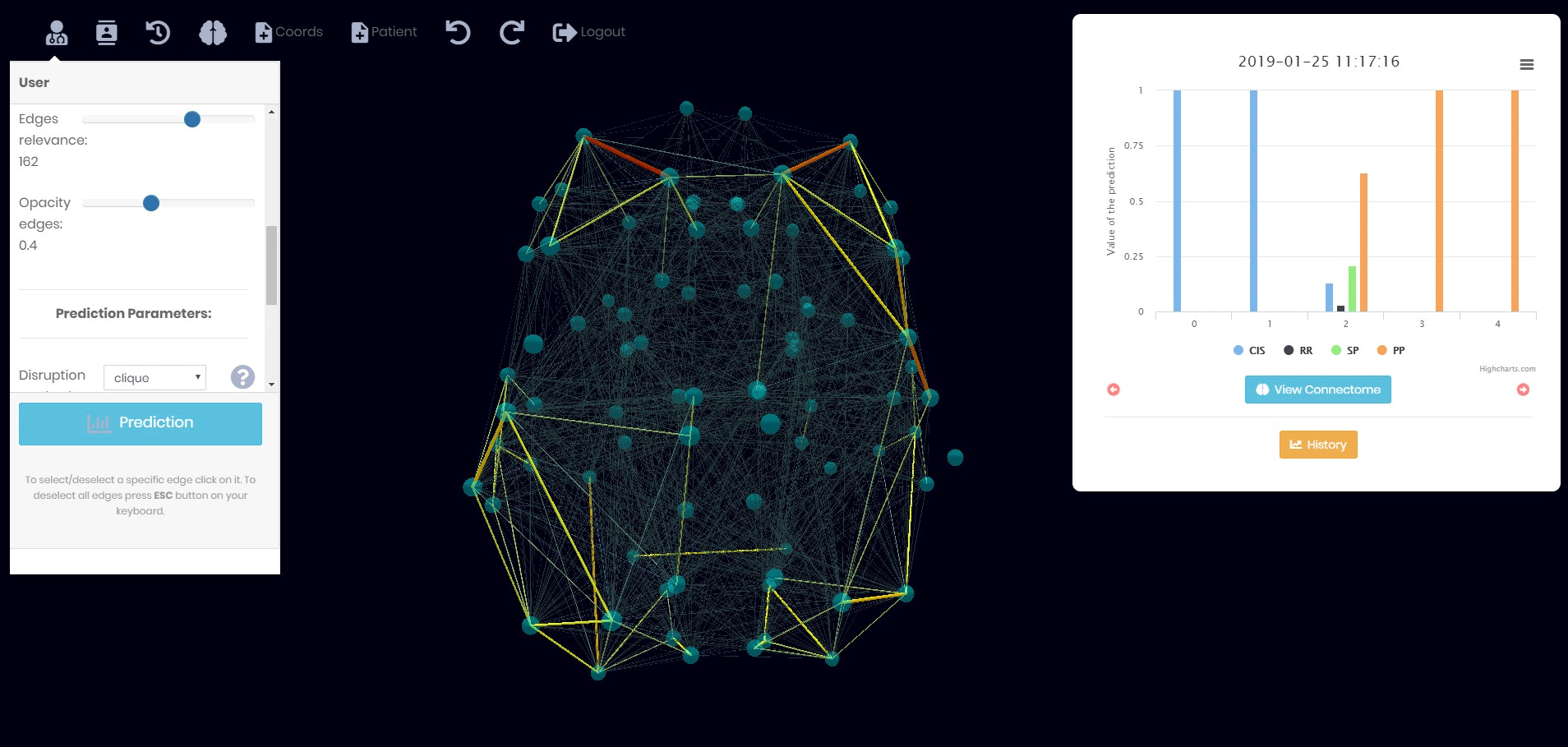}
	\caption{Screenshot of the integrated web environment\label{fig:integrated_web_environment}}
\end{figure}

The input to the tool is expected to be a graph representation of the brain, obtained as described in Section~\ref{sub:mri-graphs}. 
A 3D environment showing the connectome is then generated, as shown in Figure~\ref{fig:integrated_web_environment}. 
The graph accurately reflects the shape of a human brain, so that it is possible to identify which nodes belong to a specific brain area. Usual rotation, translation and zooming operations are available for inspecting the brain structure. 
Edge colors depend on the corresponding weight, so as to provide a visual representation of connection strength.

The tool provides pre-defined specializations for the various modules of the framework. 
For some of them, such as the \emph{Brain Evolution Simulation} module, the user can choose one among different ASP programs already available, or she/he can provide other programs personally developed for specific purposes. 
It allows also to choose an empty program, in order to apply only manual modifications as explained below.

By launching the classification and asking for one single iteration of the framework, the user can immediately check (see Figure~\ref{fig:integrated_web_environment}) the new prediction on the right panel, and the 3D brain representation is updated with the applied modifications.
If the number of required iterations is more than one, the right panel shows the graphs for each iteration (similarly to the ones presented in Section~\ref{sec:experiments}), whereas only the last 3D brain representation is shown.

\begin{figure}[h]
	\centering \includegraphics[width=.8\textwidth]{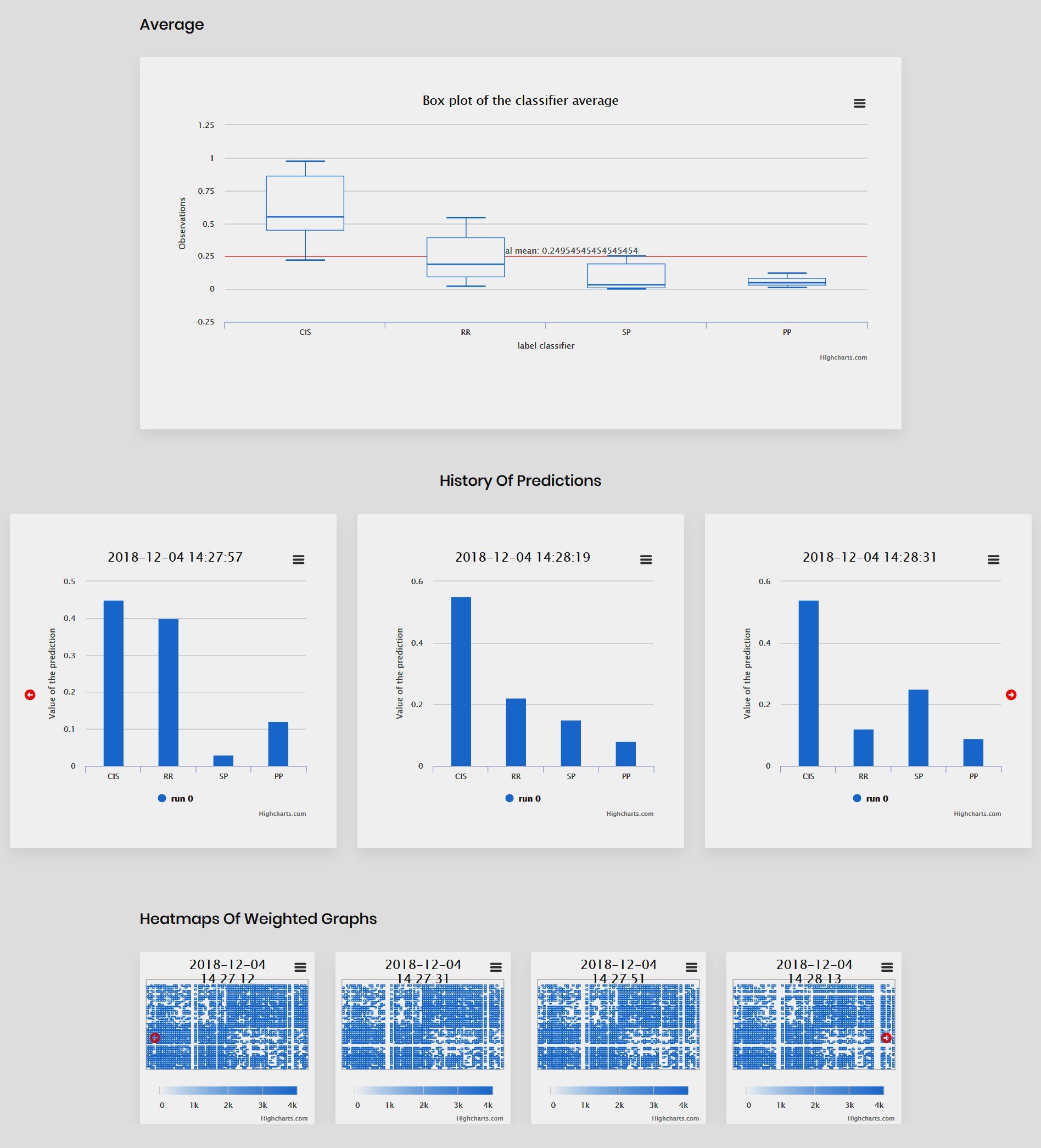}
	\caption{An example of how the environment encompasses results.\label{fig:results_page}}
\end{figure}

Once the user carried out several runs, it is possible to have a general overview of obtained results by clicking the \emph{History} button. 
In this case, the page shown in Figure~\ref{fig:results_page} is presented; it first shows a \emph{boxplot} for each stage of MS that summarizes the overall probability values returned by the classifier during the current test session.
Moreover, the detailed history of prediction results computed on the current connectome is also provided. 
Finally, for each prediction, the system provides also a {\color{\red}heat map} representing the adjacency matrix of the corresponding graph; this can be useful to see distribution of edges and weights at a glance.

Besides the implementation of the framework, the tool provides also some more functionalities helping experts to carry out manual and more refined analyses. 
Specifically, first of all, the user can manually select the set of edges to modify in a brain evolution step. 
This selection can be either exploited in substitution of the Brain Evolution Simulation module (if the user chooses the empty ASP program) or it can be seen as a pre-processing step on the connectome, if one of the specializations of the Brain Evolution Simulation module is chosen.

Furthermore, the user can modify the visualization of the connectome, based on edge importance. 
In particular, if at least one classification has been carried out on the connectome, edge importance, as introduced in Section~\ref{sub:insights-spec}, is available.
Then, users can hide or show edges, based on their importance, by using a slider.
As a consequence, manual inspection on the connectome can be greatly simplified, allowing the user to concentrate her/his attention on important edges only.

\section{Specialization of the framework to other scenarios}
\label{sec:furtherapplication}

{\color{\red}
In order to show the generality of the proposed framework, in this section we present some additional application scenarios it can be specialized to in a quite straightforward way. In particular, we first overview some additional biomedical contexts and then we show the application of the framework to a very different application scenario, namely  influence prediction in social networks.

\subsection{Specialization of the framework to other neurological disorders}
It has been proved by several independent studies that there is a strong correlation between the variations of connections among neurons and the function of the brain and, consequently, with possible insurgence of several neurological disorders~\cite{BaMa13}.
As an example, in the Alzheimer Disease researchers observed a decrease in the connectivity, associated with changes in the hippocampus~\cite{Le*15}; altered connectivity has been observed also in the Parkinson disease~\cite{Le*15}; similarly, an increased connectivity associated with changes in the amygdala have been associated with  anxiety disorder~\cite{Stein*07}. 
In all such contexts, the analysis of brain connections and their variations can provide significant insights in the knowledge of disease evolution.

Let us concentrate on the Alzheimer Disease (hereafter, AD); it is well known that, at early stages, this disease appears as Mild Cognitive Impairment (hereafter, MCI) but not all patients with MCI subsequently develop AD~\cite{Petersen04}. 
However, it is also known that in MCI progression towards AD a key role is played by the loss of connectivity among the different cortical areas. 
Thus, a large variety of approaches aiming at characterizing both MCI and AD have been proposed in the literature~\cite{Hornero*09,Jeong04}.
Some of them are based on the analysis of electroencephalograms (EEG) data; this is a less invasive observation method than MRI.
In this case, the analysis can be based on a graph, where each node represents an electrode, and the weight of each edge expresses the similarity degree between the signals registered by the corresponding electrodes (see ~\cite{Terracina-JDMMM}).

Our framework can be specialized quite straightforwardly to the analysis of MCI evolution in order to study key factors determining its progress to AD. 
In fact, it is sufficient to specialize the \emph{Classifier} module with any approach developed to classify patients in one of the  {\em Healthy}, {\em MCI}, or {\em AD} stages. 
The \emph{Brain Evolution Simulation} module can be then specialized with a proper ASP program that identifies subgraphs corresponding to (variations of) some property of interest; as an example, in~\cite{Terracina-JDMMM} it has been shown that interesting graph properties  related to AD are network density and clustering coefficient. 
Here, again, the \emph{Classification Validity Check} can refute the classification outcome if a significant degradation in the graph corresponds to a predicted remission of the disease; indeed, this situation is not biologically relevant.

\subsection{Specialization of the framework in the context of Social Networks}

The work presented in~\cite{Wu*19} surveys several contexts where ANNs are applied to graph data; indeed, there is increasing interest in extending deep learning approaches for this kind of data.
Interesting applications include, but are not limited to, e-commerce and recommender systems, citation networks, social networks, traffic analysis, drug discovery, adversarial attack prevention, and event detection. 
All of these problems can  benefit from the application of our framework in the identification of potentially relevant graph properties.

In order to provide an example, we focus next on one of them, namely influence prediction in social networks~\cite{Qiu*18}. 
A social network can be represented as a graph $G=(V,E)$, where $V$ denotes the set of users and $E$ denotes the set of relationships between them. 
Each user in a social network performs social actions towards other users; these can be suitably summarized as edge labels between the nodes corresponding to the involved  users.  
Social influence commonly refers to the phenomenon that the opinions and actions of a user are affected by others. 
In many applications, such as advertising and recommendation, it is crucial to predict the social influence of each user.

In~\cite{Qiu*18} a neural network-based approach is proposed to predict the action status of a user given the action statuses of both near neighbors and local structural information. 
The list of action statuses strictly depends on the kind of social network under analysis. 
As an example it can be a ``retweet'' action in \emph{Twitter} or a citation action in academic social networks. 
The input network is then fed to the ANN which outputs a two-dimension representation for each user indicating the action status prediction, which is then exploited for the social influence computation.

In this contexts it would be of great relevance to study the evolution of social influence with respect to modifications on the social graph. 
As an example, it would be interesting to find the minimal graph modifications required to increase the social influence in the network.  
Our framework can be specialized quite straightforwardly also to this context. 
In particular, the \emph{Classifier} module can be specialized with the approach presented in~\cite{Qiu*18} in order to predict action statuses and, thus, social influence. 
The \emph{Brain Evolution Simulation} module, which in this context could be more appropriately renamed as \emph{Graph Evolution Simulation}, can be specialized with the ASP program of choice identifying the minimal changes to be applied on the input graph in order to reach some target value of social influence. 
Even if the complete development of this specific application is out of the scope of this paper, it is easy to see that the \emph{Classification Validity Checker} can be easily encoded with specific rules allowing to detect wrong classifications in action statuses.

} 

\section{Conclusion}
\label{sec:conclusion}

{\color{\red}
This paper introduced a general and extensible framework pointing out opportunities provided by a combined use of ASP and ANN. In particular, we grounded the framework in order to provide an effective support for neurologists in studying the evolution of neurological disorders.

We have shown that a mixed use of ASP and ANNs can be of significant impact both in bioinformatics and other research fields.
Indeed, logic-based modules greatly simplify the exploration of different, possibly complex, variations in the structure of the connectome, and ANNs allow to immediately check the potential impact of such variations on the course of the disease.}
We provided three specializations of the general framework and tested them on real data to show the effectiveness of the proposed approach.
{\color{\red} Extensive tests proved the potential impact of the framework on the discovery process and some limitations of current ASP solvers.}
Based on this experience, we developed a web tool allowing even non experts to explore the connectome and test the impact of its variations on the course of the disease.

{\color{\red}We believe that the results are encouraging; moreover, they further motivate the already running research activities for optimizing ASP program evaluations. Finally, obtained results provide us with a solid basis for encouraging the communities of both ASP and ANN areas to identify more contexts where a mixed use of these tools can lead to significant benefits.

As far as future work is concerned, we plan to specialize the presented framework on the application contexts outlined in Section~\ref{sec:furtherapplication}.}

\section{Acknowledgements}\label{sec:ack}
{\color{\red}
The authors would like to thank the Editors for their comments and support during the review process, and the anonymous reviewers for their helpful and constructive comments that greatly helped at improving the final version of the paper.

\smallskip
This work was partially supported by:
\textit{(i)} the Italian Ministry for Economic Development (MISE) under the project ``Smarter Solutions in the Big Data World'', funded within the call ``HORIZON2020'' PON I\&C 2014-2020 (CUP B28I17000250008); and \textit{(ii)} the Italian Ministry of Education, Universities and Research (MIUR) and the Presidency of the Council of Ministers under project ``Declarative Reasoning over Streams'' (CUP H24I17000080001) within the call \emph{Progetti di ricerca di Rilevante Interesse Nazionale} ``PRIN'' 2017, project code 2017M9C25L\_001.
}



\bibliographystyle{acmtrans}
\bibliography{bibliography}

\end{document}